\documentclass{article}

\PassOptionsToPackage{numbers, compress}{natbib}

\usepackage{markdown}
    \usepackage[preprint]{neurips_2025}

\usepackage[utf8]{inputenc} 
\usepackage[T1]{fontenc}    
\usepackage{xcolor}         
\definecolor{C1}{HTML}{660874}
\usepackage[colorlinks,urlcolor=C1]{hyperref}       
\usepackage{url}            
\usepackage{booktabs}       
\usepackage{amsfonts}       
\usepackage{nicefrac}       
\usepackage{microtype}      
\usepackage{subfigure}

\usepackage{amsthm}

\newtheorem{proposition}{Proposition}[section]

\newtheorem{definition}{Definition}[section]

\usepackage{cite}
\usepackage{enumerate}
\usepackage{graphics} 
\usepackage{epsfig} 
\usepackage{mathptmx} 
\usepackage{times} 
\usepackage{amsmath} 
\usepackage{dutchcal}
\usepackage{makecell}

\usepackage{cleveref}
\usepackage{amssymb}  
\usepackage{mathtools}
\usepackage{bbm}
\usepackage{mathrsfs}

\usepackage{algpseudocode}
\usepackage[linesnumbered, ruled]{algorithm2e}
\SetKwRepeat{Do}{do}{while}%
\usepackage{algpseudocode}
\usepackage{listings}
\usepackage{hyperref} 
\usepackage{booktabs} 
\usepackage{graphicx}
\usepackage{subfigure}
\usepackage{bbm}
\usepackage{enumitem}
\usepackage{amsmath,amssymb,amsthm}
\usepackage[export]{adjustbox}
\usepackage{changes}
\usepackage{multirow}
\usepackage[most]{tcolorbox}
\usepackage{empheq}
\usepackage{makecell}
\newcommand\given[1][]{\:#1\vert\:}
\newtcbox{\mybox}[1][blue]
  {on line, arc = 0pt, outer arc = 0pt,
    colback = #1!30!white, colframe = #1!60!black,
    boxsep = 0pt, left = 1pt, right = 1pt, top = 2pt, bottom = 2pt,
    boxrule = 0pt, bottomrule = 1pt, toprule = 1pt}

\title{Reinforcement Fine-Tuning Powers Reasoning Capability of Multimodal Large Language Models}

\author{%
  Haoyuan Sun, Jiaqi Wu, Bo Xia, Yifu Luo, Yifei Zhao, Kai Qin, Xufei Lv, \\\textbf{Tiantian Zhang, Yongzhe Chang, Xueqian Wang} \\Tsinghua Shenzhen International Graduate School, Tsinghua University\\
    \texttt{sun-hy23@mails.tsinghua.edu.cn}\\
 {\color{C1}Project: https://github.com/Sun-Haoyuan23/Awesome-RL-based-Reasoning-MLLMs}
}

\begin{document}
\maketitle
\begin{abstract}
Standing in 2025, at a critical juncture in the pursuit of Artificial General Intelligence (AGI), reinforcement fine-tuning (RFT) has demonstrated significant potential in enhancing the reasoning capability of large language models (LLMs) and has led to the development of cutting-edge AI models such as OpenAI-o1 and DeepSeek-R1. Moreover, the efficient application of RFT to enhance the reasoning capability of multimodal large language models (MLLMs) has attracted widespread attention from the community. In this position paper, we argue that reinforcement fine-tuning powers the reasoning capability of multimodal large language models. To begin with, we provide a detailed introduction to the fundamental background knowledge that researchers interested in this field should be familiar with. Furthermore, we meticulously summarize the improvements of RFT in powering reasoning capability of MLLMs into five key points: diverse modalities, diverse tasks and domains, better training algorithms, abundant benchmarks and thriving engineering frameworks. Finally, we propose five promising directions for future research that the community might consider. We hope that this position paper will provide valuable insights to the community at this pivotal stage in the advancement toward AGI. 
Summary of works done on RFT for MLLMs is available at \href{https://github.com/Sun-Haoyuan23/Awesome-RL-based-Reasoning-MLLMs}{the project}.
\end{abstract}

\section{Introduction}
\label{Introduction}
\begin{center}
\textit{``The senses are the organs by which man perceives the world, and the soul acts through them as through tools.'' \qquad \qquad \qquad \qquad \qquad \qquad \qquad \qquad \qquad \qquad \qquad \quad —— Leonardo da Vinci}
\end{center}

Reinforcement learning (RL), a series of machine learning approaches in which an agent learns optimal decision-making strategies by continuously employing the trial-and-error paradigm \citep{sutton1998reinforcement}. Over the past four decades, from classic algorithms to deep neural networks, from value-based to policy-based methods, this fascinating field has witnessed consistent and substantial advancements through dedicated research and exploration. Currently at the threshold of 2025, Proximal Policy Optimization (PPO) \citep{schulman2017proximal} stands out as one of the most influential RL algorithms within the community.

Since the 2020s, the emergence of large language models (LLMs) has rapidly accelerated advancements across numerous interdisciplinary fields. Their remarkable zero-shot capabilities, along with emerging reasoning and planning abilities, have offered a glimpse of the potential for achieving Artificial General Intelligence (AGI). The reinforcement learning from human feedback (RLHF) \citep{ouyang2022training} pipeline has facilitated the development of epoch-making models, exemplified by GPT-4 \citep{achiam2023gpt} and LLaMA \citep{touvron2023llama,touvron2023llama2}. However, their intellectual capabilities are largely constrained by human annotators. This limitation is particularly evident in the models' reasoning abilities, specifically in their systematic capacity to derive logical inferences, methodically solve complex problems, and effectively transfer knowledge across diverse domains. Subsequently, OpenAI-o1 \citep{openaio1} successfully applied large-scale reinforcement learning to the model training, which has significantly enhanced its reasoning capabilities. Moreover, DeepSeek-R1-Zero \citep{guo2025deepseek} has demonstrated remarkable self-evolution capabilities through a pure reinforcement learning process; additionally, DeepSeek-R1 \citep{guo2025deepseek} further stabilizes the RL process through cold start, ultimately exhibiting unparalleled reasoning capabilities. Their success has demonstrated the effectiveness of Reinforcement Fine-Tuning (RFT) in enhancing the reasoning capabilities of LLMs. However, it is noteworthy that their reasoning process involves only the textual modality.
\begin{figure}[t]
\centering
\includegraphics[width=1\linewidth]{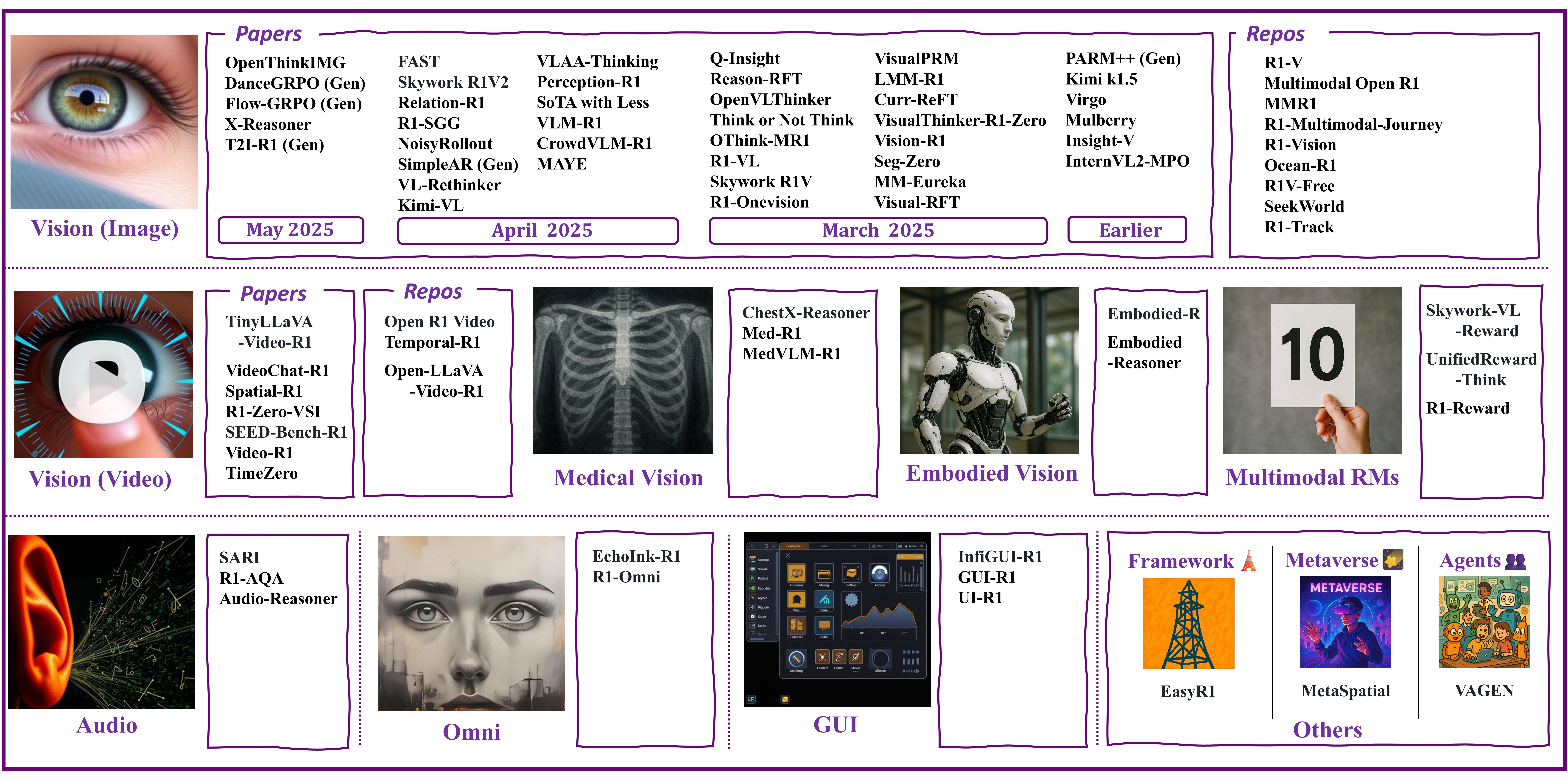}
\caption{An overview of works done on reinforcement fine-tuning (RFT) for multimodal large language models (MLLMs). Works are sorted by release time and are collected up to May 15, 2025. Further detailed summary is provided in \Cref{Summary of works done on RFT for MLLMs}.}
\label{multimodal}
\end{figure}

Just as human perception depends on the harmonious interplay of multiple senses to form a coherent understanding of the world, multimodal large language models (MLLMs) integrate diverse data modalities (such as vision, text, audio, and so on) to perceive and reason about the complex and multimodal environments. Within this context, RFT has served as a critical mechanism for powering MLLMs with robust reasoning capabilities. With the emergence of DeepSeek-R1 \citep{guo2025deepseek}, the efficient application of such training paradigm to enhance MLLM reasoning capabilities has attracted widespread attention from the community. A summary of these recent works is provided in \Cref{multimodal} and \Cref{Summary of works done on RFT for MLLMs}. They have demonstrated that RFT significantly enhances the reasoning abilities of MLLMs, making them more proficient across diverse modalities, tasks, and domains. Our position is:
\begin{tcolorbox}[enhanced,colback=white,%
    colframe=C1!75!black, attach boxed title to top right={yshift=-\tcboxedtitleheight/3, xshift=-.75cm}, title=\large{\textbf{POSITION}}, coltitle=C1!75!black, boxed title style={size=small,colback=white,opacityback=1, opacityframe=0}, size=title, enlarge top initially by=-\tcboxedtitleheight/2]

\textcolor{C1!}{\textit{\textbf{Reinforcement Fine-Tuning (RFT) Powers Reasoning
Capability of Multimodal Large Language Models (MLLMs).}}
}
\end{tcolorbox}

Standing at the pivotal moment of 2025, we believe that the field of MLLM reasoning is experiencing an exciting period of transformation. Despite the challenges faced, this era presents unique opportunities for significant advancements. This position paper aims to provide the community with a valuable reference. Specifically, we are dedicated to answering the following three questions: 

1. \textbf{\textit{What background should researchers interested in this field know?}} We answer this question in \Cref{Backgrounds}, which can be divided into three parts. To enhance readers' understanding of this topic, we begin with the fundamental concepts and classic algorithms of {\color{C1}reinforcement learning} in \Cref{Reinforcement Learning}, which are generally categorized into value-based and policy-based methods. Furthermore, following the recent survey \citep{lin2025mind}, we present a concise overview of the current state of {\color{C1}multimodal reasoning} in \Cref{Multimodal Reasoning}, showing the trend from language-centric multimodal reasoning to collaborative multimodal reasoning. Finally, in \Cref{Reinforcement Fine-Tuning}, we provide a comprehensive discussion of {\color{C1}representative RFT algorithms}, emphasizing their similarities and differences; moreover, partly drawing on the recent survey \citep{zhou2025reinforced}, we categorize them as Critic-Model-Driven and Critic-Model-Free algorithms. 

2. \textbf{\textit{What has the community done?}} We meticulously answer this question from five perspectives, as outlined in \Cref{RFT for MLLMs: What has the community done?}. Firstly, the community has made significant process in the reasoning of {\color{C1}diverse modalities}. Secondly, significant progress has also been achieved across {\color{C1}diverse tasks and domains}. Thirdly, the community has advanced the development of {\color{C1}better training algorithms}. Fourthly, we reveal several exciting trends in the development of {\color{C1}abundant reasoning multimodal benchmarks}. Finally, {\color{C1}thriving engineering frameworks} have been continuously developed by the community.

3. \textbf{\textit{What could the community do next?}} We critically answer this question by presenting five key points, as demonstrated in \Cref{Future Work: What could the community do next?}. Firstly, the community could further explore strategies for {\color{C1}enhancing generalization across various modalities, tasks, and domains}. Secondly, effectively {\color{C1}combining the strengths of the outcome reward paradigm and the process reward paradigm} is a promising direction for further research. Thirdly, we hope the community could devote increased attention to the {\color{C1}safety of reasoning MLLMs}. Fourthly, given the scarcity of multimodal data, further research into {\color{C1}data augmentation techniques} for reasoning scenarios is highly promising. Finally, the community could continue to conduct in-depth research on {\color{C1}more effective algorithms, improved reward paradigms, and broader applications}.

\section{Background}
\label{Backgrounds}
\subsection{Reinforcement Learning: Value-based and Policy-based Methods}
\label{Reinforcement Learning}
\textbf{Markov Decision Process.} The Markov Decision Process (MDP) \citep{bellman1957markovian, sutton1998reinforcement} is a foundational mathematical framework for modeling sequential decision-making in stochastic environments. It extends Markov chains by incorporating actions and rewards, enabling agents to learn optimal policies that maximize cumulative long-term rewards. Formally, an MDP is defined by a tuple $(\mathcal{S}, \mathcal{A}, \mathcal{P}, \mathcal{R}, \mathcal{\rho}, \mathcal{\gamma})$. Herein, $\mathcal{S}$ represents the state space; $\mathcal{A}$ donates the action space; and $\mathcal{P}$ is the state transition function. At any time step $t$, for the state $s_t, s_{t+1} \in \mathcal{S}$ and the action $a_t \in \mathcal{A}$, $\mathcal{P}(s_{t+1} | s_t, a_t)$ denotes the probability of reaching state $s_{t+1}$ after performing the action $a_t$ in state $s_{t}$. $\mathcal{R}$ is the reward function, where $\mathcal{R}(s_t, a_t) \in \mathbb{R}$. $\mathcal{\rho}$ denotes the initial state distribution; and $\mathcal{\gamma}$ serves as the discount factor, indicating the importance of future rewards in relation to the current state. The agent-environment interaction typically operates in a blocking paradigm. At step $t$, the agent observes state $s_t$ and, based on its policy $\pi(\cdot|s_t)$, performs action $a_t$. After taking the action, the environment updates to a new state $s_{t+1} \sim \mathcal{P}(\cdot | s_t, a_t)$, and yielding a reward $\mathcal{R}(s_t, a_t)$ for the agent.
Reinforcement learning aims to acquire an optimal policy, denoted as $\pi^*(\cdot|s_t)$, by maximizing the cumulative discounted reward (also known as the return), $G_t = \sum_{k=t}^{T} \gamma^{k-t} \mathcal{R}(s_k,a_k)$, where $T$ represents the episode horizon.  

\textbf{Classic Reinforcement Learning.} Since the 1980s, prominent computer scientists Andrew G. Barto and Richard S. Sutton have laid theoretical foundations and developed key algorithms in the field of reinforcement learning. Recently, the 2024 ACM A.M. Turing Award was conferred upon them in recognition of their fundamental contributions to the field of reinforcement learning. Furthermore, their book ``\textit{Reinforcement Learning:
An Introduction}'' \citep{sutton1998reinforcement} has earned the title of ``Bible of Reinforcement Learning'' among researchers. Hence, we argue that a fundamental understanding of classic reinforcement learning is crucial for practitioners employing reinforcement fine-tuning techniques, especially to avoid ambiguity in conceptual definitions. We restate the definitions:
\begin{definition}[State Value Function \citep{sutton1998reinforcement}]
\textbf{The value of a state $s$ under a policy $\pi$}, denoted $V_\pi(s)$, is the expected return
when starting in $s$ and following $\pi$ thereafter. For MDPs, $V_\pi(s)$ can be defined as:
\begin{equation}
V_\pi(s)=\mathbb{E}_{\pi}\mybox[violet]{\text{$[G_t\given S_t=s]$}}=\mathbb{E}_{\pi}\Bigg[\sum_{k=0}^{\infty} \gamma^k\mathcal{R}_{t+k+1} \given[\Bigg] S_t=s \Bigg].
\end{equation}
\end{definition}
\begin{definition}[Action Value Function \citep{sutton1998reinforcement}]
\textbf{The value of taking action {$a$} in state $s$ under policy $\pi$}, denoted $Q_\pi(s,a)$, as expected return starting from $s$, taking action
$a$, thereafter following policy $\pi$:
\begin{equation}
Q_\pi(s, a)=\mathbb{E}_{\pi}\mybox[violet]{\text{$[G_t\given S_t=s, A_t=a]$}}=\mathbb{E}_{\pi}\Bigg[\sum_{k=0}^{\infty} \gamma^k\mathcal{R}_{t+k+1} \given[\Bigg] S_t=s, A_t=a \Bigg].
\end{equation}
\end{definition}
{\color{C1}Classic reinforcement learning algorithms} are categorized as {\color{C1}\uline{Value-based}} methods and {\color{C1}\uline{Policy-based}} methods. {\color{C1}\uline{Value-based}} methods primarily focus on learning a value function (usually the action value function $Q_\pi(s, a)$), from which a policy is subsequently derived. Early typical algorithms within such paradigm include Q-Learning \citep{sutton1998reinforcement, melo2001convergence} and SARSA \citep{sutton1998reinforcement}. With the rise of deep learning, researchers have increasingly employed neural networks to approximate the value function, leading to a series of improvements, exemplified by Deep Q-Network (DQN) \citep{mnih2013playing, mnih2015human}, Double DQN \citep{van2016deep}, Dueling DQN \citep{wang2016dueling}, Rainbow \citep{hessel2018rainbow}, and so on. In contrast, {\color{C1}\uline{Policy-based}} methods directly and explicitly learn a target policy with the primary objective of identifying an optimal policy that maximizes the expected reward within the environment. REINFORCE \citep{sutton1999policy} pioneers this paradigm, defining the target function as $J(\theta)=\mathbb{E}_{s_0}[V_{\pi_{\theta}}(s_0)]$, where $s_0$ is the inital state; subsequently, the objective function is differentiated with respect to the policy parameter $\theta$, and gradient ascent method is applied to maximize this function. Actor-Critic algorithms \citep{sutton1998reinforcement, konda1999actor, mnih2016asynchronous} take a further step by fitting a value function to guide policy learning.
Critic (value module), learns to discriminate between effective and ineffective actions based on data sampled by Actor (policy module), subsequently guiding Actor's policy update; concurrently, as Actor trains, the distribution of environment interaction data shifts, necessitating rapid adaptation for Critic to provide accurate value estimations under the evolving data distribution. {\color{C1}It's important to clarify that Actor-Critic family algorithms are essentially \uline{Policy-based} algorithms}: as they fundamentally aim to optimize the policy while only concurrently learning a value function to enhance policy learning efficiency.

\textbf{From TRPO to PPO. } Trust Region Policy Optimization (TRPO) \citep{schulman2015trust} and Proximal Policy Optimization (PPO) \citep{schulman2017proximal} inherit the Actor-Critic paradigm; therefore, they are actually {\color{C1}\uline{Policy-based}} methods. A significant limitation of previous {\color{C1}\uline{Policy-based}} methods is the potential for abrupt policy degradation when updating parameters along the policy gradient (often attributed to excessively large step sizes). TRPO promotes stable and effective policy learning by employing a ``trust region'' constraint during policy updates; and it theoretically guarantees monotonic improvement in policy learning. We present the theoretical policy optimization objective of TRPO in \Cref{TRPO} without any further proof.
\begin{proposition}[TRPO Objective \citep{schulman2015trust}]
\label{TRPO}
The current policy, parameterized by $\theta_{\text{old}}$, is denoted as $\pi_{\theta_{\text{old}}}$. Primary goal is to find a better policy $\pi_{\theta}$ utilizing current policy $\pi_{\theta_{\text{old}}}$. The objective is:
\begin{equation}
\max_{\theta} \mathbb{E}_{s\sim\upsilon^{\pi_{\theta_{\text{old}}}}} \mathbb{E}_{a\sim\pi_{\theta_{\text{old}}}}\Bigg[ \frac{\pi_{\theta}(a|s)}{\pi_{\theta_{\text{old}}}(a|s)}A_{\pi_{\theta_{\text{old}}}}(s,a)\Bigg] \quad \text{s.t.} \;\mathbb{E}_{s\sim\upsilon^{\pi_{\theta_{\text{old}}}}}\Big[\mathbb{D}_{\text{KL}}(\pi_{\theta_{\text{old}}}(\cdot|s),\pi_{\theta}(\cdot|s))\Big]\leq \delta,
\end{equation}
where $\upsilon^{\pi}$ is the state visitation distribution under policy $\pi$; $A_\pi(s,a)$ is the advantage function with definition of $A_\pi(s,a)=Q_{\pi}(s,a)-V_{\pi}(s)$; $\mathbb{D}_{\text{KL}}$ is the Kullback-Leibler divergence that serves as a constraint to ensure that the new policy remains sufficiently close to the old policy.
\end{proposition}
However, TRPO practically solves the objective using methods such as Taylor expansion approximation, conjugate gradient, and line search, making computational cost of each update step significantly high. Thereby, the Proximal Policy Optimization (PPO) \citep{schulman2017proximal} is proposed, directly incorporating first-order optimization through a clipped objective function (PPO-Clip) or a KL divergence penalty term (PPO-Penalty), avoiding second-order calculation. PPO objectives are presented in \Cref{PPO}.
\begin{proposition}[PPO Objectives \citep{schulman2017proximal}]
\label{PPO}
PPO-Penalty incorporates KL divergence constraint into the objective function using Lagrangian multiplier method; and coefficient $\beta$ is updated during training:
\begin{equation}
\max_{\theta}\mathbb{E}_{s\sim\upsilon^{\pi_{\theta_{\text{old}}}}} \mathbb{E}_{a\sim\pi_{\theta_{\text{old}}}}\Bigg[ \frac{\pi_{\theta}(a|s)}{\pi_{\theta_{\text{old}}}(a|s)}A_{\pi_{\theta_{\text{old}}}}(s,a)-\mybox[violet]{\text{$\beta\mathbb{D}_{\mathrm{KL}}(\pi_{\theta_{\text{old}}}(\cdot|s),\pi_{\theta}(\cdot|s))$}} \Bigg]
\end{equation}
Setting $d=\mathbb{E}_{s\sim\upsilon^{\pi_{\theta_{\text{old}}}}}[\mathbb{D}_{\text{KL}}(\pi_{\theta_{\text{old}}}(\cdot|s),\pi_{\theta}(\cdot|s))]$. If $d<d_{\text{targ}}/1.5$, $\beta\leftarrow\beta/2$; if $d>d_{\text{targ}}\times1.5$, $\beta\leftarrow2\beta$.
PPO-Clip restricts the objective function more directly with the clip operation: 
\begin{equation}
\label{PPO_clip}\max_{\theta}\mathbb{E}_{s\sim\upsilon^{\pi_{\theta_{\text{old}}}}} \mathbb{E}_{a\sim\pi_{\theta_{\text{old}}}}\Bigg[ \mybox[violet]{\text{$\min$}}\Bigg( \frac{\pi_{\theta}(a|s)}{\pi_{\theta_{\text{old}}}(a|s)}A_{\pi_{\theta_{\text{old}}}}(s,a), \mathrm{\mybox[violet]{\text{$\mathrm{clip}$}}}\Bigg( \frac{\pi_{\theta}(a|s)}{\pi_{\theta_{\text{old}}}(a|s)}, \mybox[violet]{\text{$1-\epsilon, 1+\epsilon$}}\Bigg)A_{\pi_{\theta_{\text{old}}}}(s,a) \Bigg) \Bigg]
\end{equation}
\end{proposition}
For PPO-Clip, as shown in \Cref{PPO_clip}, if the advantage function is positive (action value is higher than average), maximizing the equation leads to an increase of $\pi_{\theta}/\pi_{\theta_{\text{old}}}$, but it is constrained not to exceed $1+\epsilon$; similarly, if the advantage function is negative (action value is lower than average), maximizing the equation leads to a decrease of $\pi_{\theta}/\pi_{\theta_{\text{old}}}$, but it is constrained not to exceed $1-\epsilon$.
\subsection{Multimodal Reasoning: Language-Centric and Collaborative Paradigms}
\label{Multimodal Reasoning}
Recent advances in the reasoning capabilities of Large Language Models (LLMs) \citep{zhang2024llm, plaat2024reasoning, xu2025towards,wang2025tutorial, bandyopadhyay2025thinking,chen2025towards, zhang2025100} pave a significant step towards achieving Artificial General Intelligence (AGI), exemplified by OpenAI-o1 \citep{openaio1}, DeepSeek-R1 \citep{guo2025deepseek}, and so on. Furthermore, sustained success of Multimodal Large Language Models (MLLMs) has motivated researchers to explore the integration of LLM reasoning capabilities with multimodal processing \citep{lin2025investigating, wang2025multimodal, lin2025mind, zhou2025reinforced,li2025perceptionreasonthinkplan}, resulting in notable implementations such as Kimi k1.5 \citep{team2025kimi}, OpenAI o3 and o4-mini \citep{openaio3}, Grok 3 \citep{grok3}, and so on. In the recent survey \citep{lin2025mind}, the authors proposed a taxonomy of multimodal reasoning into two categories: {\color{C1}{\uline{the Language-Centric Multimodal Reasoning}}} and {\color{C1}{\uline{the Collaborative Multimodal Reasoning}}}. We argue that such taxonomy is insightful and constructive; furthermore, such taxonomy effectively illustrates an evolution in multimodal reasoning technologies, specifically a shift from language-dominated control to collaborative multimodal co-reasoning.
Hence, we further follow this taxonomy \citep{lin2025mind}:

\textbf{Language-Centric Multimodal Reasoning \citep{lin2025mind}. }In this paradigm, the multimodality (beyond language) primarily functions to acquire perceptual information and extract features; the reasoning process, on the contrary, is predominantly driven by the language modality. This paradigm is further divided into the One-pass Multimodality Perception and the Active Multimodality Perception based on the multimodal perception triggering mechanisms. One-pass Multimodality Perception methods treat multimodal information (beyond language) as static context, encoding them only once during the model’s input stage. In contrast, for Active Multimodality Perception methods, language modality's generation of intermediate reasoning steps triggers iterative multimodal re-perception cycles. 

\textbf{Collaborative Multimodal Reasoning \citep{lin2025mind}. }In this paradigm, the reasoning process further necessitates multimodal (beyond language) action reasoning and multimodal (beyond language) state updating, and multimodal representation extends beyond passive perception to active collaboration with the language modality throughout the reasoning process. Regarding the multimodal (beyond language) action reasoning, it transcends purely linguistic instructions and generates internal reasoning actions autonomously; furthermore, it exhibits explicit reasoning trajectories within the multimodal feature space. Then, the model dynamically updates multimodal contextual information by executing the aforementioned multimodal (beyond language) actions, which actually introduce new constraints to the language modality and thereby trigger subsequent multimodal reasoning steps.

\subsection{Reinforcement Fine-Tuning: Critic-Model-Driven and Critic-Model-Free Algorithms}
\label{Reinforcement Fine-Tuning}
Leveraging reinforcement fine-tuning (RFT), recent studies have introduced novel post-training algorithms to enhance reasoning capabilities of LLMs \citep{zhang2025100,openaio1,guo2025deepseek} and MLLMs \citep{team2025kimi, openaio3, grok3}. In the recent survey \citep{zhou2025reinforced}, the authors categorized RL-based training into value-based and value-free algorithms. While this specific point is correct within the RFT context, we argue that it could conflict with the concepts of classic reinforcement learning (for example, PPO is classified as a value-based algorithm within this system; however, it is actually a {\color{C1}\uline{Policy-based}} algorithm). Herein, we refine the taxonomy for {\color{C1}reinforcement fine-tuning} as: {\color{C1}\uline{Critic-Model-Driven}} algorithms and {\color{C1}\uline{Critic-Model-Free}} algorithms.

\textbf{Critic-Model-Driven algorithms. }Since the introduction of PPO \citep{schulman2017proximal} in 2017, it has become one of the most popular actor-critic RL algorithms for policy optimization of LLMs \citep{ouyang2022training,hu2025open} and MLLMs. Within the context of MLLMs, the input $(m,t)$ consisting of both multimodal (beyond language) contents $m$ and a textual query $t$; then, PPO objective in \Cref{PPO_clip} can be transferred as \citep{zhang2025100,hu2025open}:
\begin{equation}
\begin{split}    \max_{\theta}\,\,\,\,\,\,\,\,&\mathbb{E}_{((m,t),a)\sim\mathcal{D},\{o_i\}_{i=1}^{G}\sim\pi_{\theta_{\text{old}}}(\cdot|(m,t))}\Bigg[\frac{1}{G}\sum_{i=1}^G \frac{1}{\left|o_i\right|}\sum_{t=1}^{\left|o_i\right|} \Bigg(\\
\min&\Bigg(\frac{\pi_{\theta}(o_{i,t}|(m,t),o_{i,<t})}{\pi_{\theta_{\text{old}}}(o_{i,t}|(m,t),o_{i,<t})} \mybox[violet]{\text{$\hat{A}_{i,t}\big(\phi\big)$}} ,\mathrm{clip}\bigg(\frac{\pi_{\theta}(o_{i,t}|(m,t),o_{i,<t})}{\pi_{\theta_{\text{old}}}(o_{i,t}|(m,t),o_{i,<t})}, 1-\epsilon, 1+\epsilon\bigg)\mybox[violet]{\text{$\hat{A}_{i,t}\big(\phi\big)$}}\Bigg) \Bigg)\Bigg],
\end{split}
\end{equation}
where $\hat{A}_{i,t}\big(\phi\big)$ represents the Generalized Advantage Estimation (GAE) \citep{schulman2015high}: it is computed {\color{C1}using the ``value'' provided by the critic model} $V_{\phi}(o_{i,t}|(m,t),o_{i,<t})$ that is trained concurrently with the policy model $\pi_{\theta}(o|(m,t))$. In the domain of LLM policy optimization, several studies have made significant improvements along this line, exemplified by Open-Reasoner-Zero \citep{hu2025open} and VC-PPO \citep{yuan2025s}.

\textbf{Critic-Model-Free algorithms. } Group Relative Policy Optimization (GRPO) \citep{guo2025deepseek,shao2024deepseekmath} {\color{C1}discards the critic model and the calculation of GAE in PPO} by sampling and normalizing rewards within a group of $G$ outputs, which significantly enhances efficiency and reduces memory consumption. In addition, a KL-divergence penalty is applied to constrain the optimized model $\pi_{\theta}(o|(m,t))$, mitigating excessive divergence from initial SFT model $\pi_{\text{ref}}(o|(m,t))$. We detail the GRPO objective in \Cref{GRPO}.

\begin{proposition}[GRPO Objective \citep{guo2025deepseek,shao2024deepseekmath}]
\label{GRPO}
For each sample $((m,t),a)$, GRPO samples a group of outputs $\{o_i\}_{i=1}^G$ from the old policy $\pi_{\theta_{\text{old}}}(o|(m,t))$, and the policy model $\pi_{\theta}(o|(m,t))$ is optimized by:
\begin{equation}
\begin{split}    \max_{\theta}\mathbb{E}_{((m,t),a)\sim\mathcal{D},\{o_i\}_{i=1}^{G}\sim\pi_{\theta_{\text{old}}}(\cdot|(m,t))}\Bigg[\frac{1}{G}\sum_{i=1}^G \frac{1}{\left|o_i\right|}\sum_{t=1}^{\left|o_i\right|} \Bigg(
\min\Bigg(&\frac{\pi_{\theta}(o_{i,t}|(m,t),o_{i,<t})}{\pi_{\theta_{\text{old}}}(o_{i,t}|(m,t),o_{i,<t})} \mybox[violet]{\text{$\hat{A}_{i,t}$}},\\
\mathrm{clip}\bigg(\frac{\pi_{\theta}(o_{i,t}|(m,t),o_{i,<t})}{\pi_{\theta_{\text{old}}}(o_{i,t}|(m,t),o_{i,<t})}, 1-\epsilon, &1+\epsilon\bigg)\mybox[violet]{\text{$\hat{A}_{i,t}$}}\Bigg) \mybox[violet]{\text{$- \beta \mathbb{D}_{\mathrm{KL}}\big(\pi_{\theta},\pi_{\mathrm{ref}}\big)_{i,t}$}}\Bigg)\Bigg],
\end{split}
\end{equation}
where $\hat{A}_{i,t} = 
\widetilde{r}_i =\Big(r\big(o_i,a\big)-\mathrm{mean}\Big(\Big\{ r\big(o_i,a\big)\Big\}_{i=1}^G\Big)\Big)/\mathrm{std}\Big(\Big\{ r\big(o_i,a\big)
 \Big\}_{i=1}^G\Big)$ is the group relative reward (advantage); and $\Big\{ r\big(o_i,a\big)\Big\}_{i=1}^G$ represents the rewards of response group $\big\{o_i\big\}_{i=1}^G$ that computed by reward function. Furthermore, the KL divergence in GRPO is calculated by the K3 estimator \citep{K3}.
\end{proposition}
For LLM policy optimization, researchers have further proposed several representative improvements towards GRPO, exemplified by DAPO \citep{yu2025dapo}, Dr.GRPO \citep{liu2025understanding}, and so on. Furthermore, considerable engineering efforts within the LLM community have focused on adapting algorithms to achieve more stable and efficient training, exemplified by Light-R1 \citep{wen2025light}, TinyZero \citep{tinyzero}, and so on. 
\section{RFT for MLLMs: What has the community done?}
\label{RFT for MLLMs: What has the community done?}
Large Reasoning Models (LRMs) represent cutting-edge AI models designed to devote more time to thinking before providing a response, thus achieving superior reasoning capabilities. With the introduction of OpenAI-o1 \citep{openaio1}, reinforcement fine-tuning has shown great potential in the domain of LLMs. However, {\color{C1}the process reward paradigm} it utilized exhibits unstable training and limited generalizability \citep{liu2025can}. Furthermore, applying this paradigm to MLLMs might be further hindered by the challenge of generalization across diverse tasks. VisualPRM \citep{wang2025visualprm} and PARM++ \citep{guo2025can} represent pioneering efforts in this area, demonstrating its great potential for future developments. Inspired by the success of DeepSeek-R1 \citep{guo2025deepseek}, the community suggests that simple rule-based rewards (i.e. {\color{C1}the outcome reward paradigm}), even without a separate learned reward model, can suffice for the autonomous development of complex reasoning capabilities in LLMs \citep{bandyopadhyay2025thinking,chen2025towards, zhang2025100}. Effectiveness of this paradigm is also rapidly and thoroughly validated within the MLLM community. In particular, since March 2025, substantial progress has been made in enhancing multimodal reasoning within this paradigm. Generally, reinforcement fine-tuning has achieved significant successes in powering the reasoning ability of MLLMs. We meticulously divide the successes into the following five points:

\textbf{\color{C1}\uline{Success 1: Diverse Modalities}. }As demonstrated in \Cref{multimodal}, recent advancements in reinforcement fine-tuning (RFT) for multimodal large language models (MLLMs) are summarized. It is indicated that RFT has significantly enhanced the reasoning abilities of vision, audio, omni-multimodal, graphical user interface (GUI), metaverse interaction and agents in MLLMs. Notably, in addition to substantial progress in the vision modality, the community has also achieved significant breakthroughs in other modalities. Audio-Reasoner \citep{xie2025audio}, R1-AQA \citep{li2025reinforcement} and SARI \citep{wen2025sari} have utilized RFT to enhance the reasoning capabilities of {\color{C1}Large Audio Language Models (LALMs)} in Audio Question Answering (AQA) tasks. R1-Omni \citep{zhao2025r1} and EchoInk-R1 \citep{xing2025echoink} have successfully implemented RFT in {\color{C1}Omni-multimodal large language models}, which fundamentally rely on both visual and auditory modalities. UI-R1 \citep{lu2025ui}, GUI-R1 \citep{luo2025gui} and InfiGUI-R1 \citep{liu2025infigui} have similarly applied RFT to advance action prediction tasks for {\color{C1}graphic user interface (GUI)} agents, thereby enhancing their understanding and control capabilities. MetaSpatial \citep{pan2025metaspatial} has made significant progress in employing RFT to enhance {\color{C1}3D spatial reasoning within metaverse scenarios}. VAGEN \citep{VAGEN} has advanced the training of {\color{C1}VLM-based visual agents} through a multi-turn RFT framework.

\textbf{\color{C1}\uline{Success 2: Diverse Tasks and Domains (Take Vision Modality as an Example)}. }As previously mentioned, RFT has achieved significant successes in diverse modalities. Furthermore, within the vision modality alone, considerable successes have been achieved across a wide range of tasks and domains. {\color{C1}Mathematical visual reasoning} \citep{lu2021inter,wang2024measuring,zhang2024mathverse,lu2023mathvista,sun2024mm,qiao2024we,zou2025dynamath} and {\color{C1}academic multi-discipline reasoning} \citep{he2024olympiadbench,yue2024mmmu,hao2025can}, tasks that receive a lot of attention from the community, require the precise integration of symbolic processing, visual analysis, and logical reasoning. Much groundbreaking works on this subject have already been carried out by the community: InternVL2-MPO \citep{wang2024enhancing}, Mulberry \citep{yao2024mulberry}, Virgo \citep{du2025virgo}, MM-EUREKA \citep{meng2025mm}, Vision-R1 \citep{huang2025vision}, LMM-R1 \citep{peng2025lmm}, VisualPRM \citep{wang2025visualprm}, MMR1 \citep{MMR1-Math2025}, R1-Onevision \citep{yang2025r1}, SkyworkR1V \citep{peng2025skywork}, R1-VL \citep{zhang2025r1}, OpenVLThinker \citep{deng2025openvlthinker}, VL-Rethinker \citep{wang2025vl}, NoisyRollout \citep{liu2025noisyrollout}, Skywork R1V2 \citep{wei2025skywork} and FAST \citep{xiao2025fast}. Meanwhile, {\color{C1}vision-driven tasks} have also attracted widespread attention from the community: VLM-R1 \citep{shen2025vlm} 
has presented evidence of the feasibility and effectiveness of applying RFT to visual understanding tasks like referring expression compression and open-vocabulary object detection; CrowdVLM-R1 \citep{wang2025crowdvlm} 
has adapted RFT to the task of crowd counting; VisualThinker-R1-Zero \citep{zhou2025r1} has employed RFT to vision-centric spatial reasoning tasks; CLS-RL and No-Thinking-RL \citep{li2025think} have utilized RFT for the few-shot image classification task; Seg-Zero \citep{liu2025seg} has applied RFT to the image segmentation task; Q-Insight \citep{li2025q} has adapted RFT to the image quality assessment task; Perception-R1 \citep{yu2025perception} has employed RFT to the visual perception task; R1-SGG \citep{chen2025compile} and Relation-R1 \citep{li2025relation} have utilized RFT for the tasks of scene graph generation and relation comprehension; R1-Track \citep{wang2025r1track} has applied RFT to the visual object tracking task; SeekWorld \citep{seekworld2025} has adapted RFT to the visual geolocation reasoning task; V-ToolRL \citep{su2025openthinkimg} has employed RFT for the adaptive invocation of external vision tools. Furthermore, a significant number of works have focused on {\color{C1}multi-task and multi-domain joint training} to simultaneously improve model performance across multiple tasks and domains: Insight-V \citep{dong2024insight}, Visual-RFT \citep{liu2025visual}, Reason-RFT \citep{tan2025reason}, ThinkLite-VL \citep{wang2025sota}, VLAA-Thinking \citep{chen2025sft}, Kimi-VL-Thinking \citep{team2025kimi-vl}, R1-Vision \citep{yu25r1vision} and Ocean-R1 \citep{ming2025oceanr1}. In the specific domain of {\color{C1}temporal vision (video)}, applications of RFT have successfully enhanced video reasoning capabilities: Open-R1-Video \citep{wang-2025-open-r1-video}, TimeZero \citep{wang2025timezero}, Temporal-R1 \citep{li2025temporalr1}, Open-LLaVA-Video-R1 \citep{Tang2025LlavaVideoR1}, Video-R1 \citep{feng2025video}, SEED-Bench-R1 \citep{chen2025exploring}, R1-Zero-VSI \citep{liao2025improved}, Spatial-R1 \citep{ouyang2025spatial}, VideoChat-R1 \citep{li2025videochat} and TinyLLaVA-Video-R1 \citep{zhang2025tinyllava}. Moreover, in {\color{C1}specific domain disciplines}, the application of RFT has also successfully enhanced the reasoning ability of domain-specific MLLMs: MedVLM-R1 \citep{pan2025medvlm}, Med-R1 \citep{lai2025med} and ChestX-Reasoner \citep{fan2025chestx} have represented significant advancements in medical reasoning capabilities for MLLMs in {\color{C1}medical vision}; Embodied-Reasoner \citep{zhang2025embodied} and Embodied-R \citep{zhao2025embodied} have demonstrated substantial reasoning capabilities progress in {\color{C1}embodied vision}. Beyond that, RFT has also further powered {\color{C1}multimodal generation (especially the text-to-image generation)}, exemplified by PARM++ \citep{guo2025can}, SimpleAR \citep{wang2025simplear}, T2I-R1 \citep{jiang2025t2i}, Flow-GRPO \citep{liu2025flow} and DanceGRPO \citep{xue2025dancegrpo}. 

\textbf{\color{C1}\uline{Success 3: Better Training Algorithms}. }In addition to exploring the application of GRPO across diverse modalities, tasks, and domains, we observe that the community has also conducted in-depth exploration into better algorithms. These explorations primarily focus on {\color{C1}\uline{training paradigm}} \citep{deng2025boosting,meng2025mm,deng2025openvlthinker,wang2025vl,liu2025noisyrollout}, {\color{C1}\uline{algorithmic strategy}} \citep{liu2025othink,zhang2025r1,xiao2025fast}, and {\color{C1}\uline{data selection}} \citep{wang2025sota}. Curr-ReFT \citep{deng2025boosting} proposes a novel post-training paradigm comprising two stages: the {\color{C1}curriculum reinforcement learning} (difficulty-aware reward design) and the {\color{C1}rejected sampling-based self-improvement} (selective learning from high-quality examples). MM-EUREKA \citep{meng2025mm} introduces {\color{C1}online filtering paradigm}, which eliminates prompts that yield responses deemed either entirely correct or entirely incorrect during training; furthermore, the study's {\color{C1}implementation of DAPO} \citep{yu2025dapo} {\color{C1}and ADORA} \citep{gui2025adora} also provides valuable insights for future improved training paradigms. OpenVLThinker \citep{deng2025openvlthinker} {\color{C1}iteratively employs SFT and GRPO}, utilizing reasoning data from previous iterations to achieve self-improvement; significantly, it evolves the training data to progressively include more challenging questions over iterations. VL-Rethinker \citep{wang2025vl} introduces the {\color{C1}Selective Sample Replay (SSR)} to mitigate the vanishing advantage issue in GRPO and incorporates the {\color{C1}Forced Rethinking} to explicitly enforce a self-reflection reasoning step. NoisyRollout \citep{liu2025noisyrollout} {\color{C1}integrates trajectories from both clean and moderately distorted images} to foster targeted diversity in visual perception and the resulting reasoning patterns; additionally, it employs a {\color{C1}noise annealing schedule} that progressively reduces distortion strength over training, maximizing the advantages of noisy signals in earlier phases while ensuring stability and scalability in later stages.
OThink-MR1 \citep{liu2025othink} introduces {\color{C1}GRPO-D}, which enhances GRPO through the incorporation of a dynamic KL divergence strategy inspired by the $\epsilon$-greedy strategy in classic reinforcement learning. R1-VL \citep{zhang2025r1} introduces {\color{C1}StepGRPO}, which incorporates both the step-wise reasoning accuracy reward and the step-wise reasoning validity reward, thereby effectively mitigating the sparse reward challenge without applying process reward models. FAST \citep{xiao2025fast} introduces {\color{C1}FAST-GRPO} that integrating three key components: model-based metrics for question characterization, an adaptive thinking reward mechanism, and difficulty-aware KL regularization. ThinkLite-VL \citep{wang2025sota} introduces an effective {\color{C1}MCTS-based data filtering method} that quantifies sample difficulty according to the number of iterations the model requires to solve each problem, thereby achieving state-of-the-art reasoning performance with fewer training samples.

\textbf{\color{C1}\uline{Success 4: Abundant Benchmarks}. }As stated in the blog \citep{Secondhalf}, abundant benchmarks have been essential on the path towards Artificial General Intelligence (AGI) in the future. In the domain of MLLM reasoning, especially in visual reasoning, there have long been general and recognized benchmarks within the community. As recent surveys \citep{lin2025investigating, wang2025multimodal, lin2025mind, zhou2025reinforced,li2025perceptionreasonthinkplan} have summarized them extensively, a detailed discussion of them is omitted here. Furthermore, our analysis reveals that following the emergence of DeepSeek-R1 \citep{guo2025deepseek}, multimodal reasoning benchmarks have shown the following exciting six trends. {\color{C1}The first trend is the increasing difficulty of benchmarks}: for example, on the ZeroBench \citep{roberts2025zerobench}, all contemporary frontier MLLMs have completely failed in this regard. {\color{C1}The second trend involves benchmarks that assess human-like reasoning capabilities}: for example, V1-33K \citep{wang2025v1} evaluates the reasoning capabilities of MLLMs by implementing auxiliary tasks, a method frequently employed in human reasoning processes; GeoSense \citep{xu2025geosense} evaluates the identification and adaptive application of geometric principles, which is an important human-like geometric reasoning mechanism that has been neglected in previous benchmarks; MM-IQ \citep{cai2025mm} evaluates the abstraction and reasoning abilities of MLLMs by utilizing human-like IQ tests. {\color{C1}The third trend is more comprehensive benchmarks on classic domains}: for example, MDK12-Bench \citep{zhou2025mdk12} extends the data size and domain coverage of the multi-discipline domain; MV-MATH \citep{wang2025mv} expands the scope of mathematical reasoning, moving from single-visual contexts to encompass multi-visual scenarios; Spatial457 \citep{wang2025pulsecheck457} innovatively broadens the scope of visual-spatial reasoning into six dimensions (6D); VCR-Bench \citep{qi2025vcr} introduces the evaluation of video chain-of-thought (CoT) reasoning for video benchmarking; MME-CoT \citep{jiang2025mme} further assesses the reasoning quality, robustness, and efficiency at a fine-grained level. {\color{C1}The fourth trend is benchmarks for more realistic application scenarios}: for example, Video-MMLU \citep{song2025video} assesses MLLMs on  multi-discipline lecture tasks; GDI-Bench \citep{li2025gdi} evaluates MLLMs on document-specific reasoning tasks. {\color{C1}The fifth trend is a transition from language-centric benchmarks to multimodal-centric (especially visual-centric) ones}: for example, VisuLogic \citep{xu2025visulogic} represents a formidable visual reasoning benchmark that inherently poses significant difficulty to articulate in language. {\color{C1}The sixth trend is the introduction of interactive elements}: for example, iVISPAR \citep{mayer2025ivispar} introduces a novel interactive benchmark designed to evaluate the spatial reasoning capabilities of VLMs acting as agents.

\textbf{\color{C1}\uline{Success 5: Thriving Engineering Frameworks}. }Within the community, enhancements to engineering training frameworks have been pivotal in reducing research barriers and increasing development efficiency. Since the emergence of DeepSeek-R1 \citep{guo2025deepseek}, several frameworks have significantly advanced the community's development. {\color{C1}Open-R1-Multimodal} has been a pioneering effort in this area that is built upon Open-R1 \citep{openr1} and TRL \citep{vonwerra2022trl}, effectively implementing multimodal model training through the GRPO algorithm. {\color{C1}R1-V} \citep{chen2025r1v} has taken a step further, making it support the Qwen2.5-VL model, the GEOQA task and the vLLM \citep{kwon2023efficient} for training acceleration. {\color{C1}EasyR1} \citep{zheng2025easyr1} is a clean fork of the original veRL \citep{sheng2024hybridflow} project. It features extensive support for models, algorithms, and datasets, along with support for padding-free training, checkpoint resumption, and tool integration. {\color{C1}MAYA} \citep{ma2025maye} offers a transparent and reproducible framework, along with a comprehensive evaluation scheme, for the application of RL to MLLMs; furthermore, it also serves as a lightweight and educational framework that elucidates the core logic of RL training.
\section{Future Work: What could the community do next?
}
\label{Future Work: What could the community do next?}
As discussed in \Cref{RFT for MLLMs: What has the community done?}, the emergence of Deepseek-R1 \citep{guo2025deepseek} has significantly increased interest within the community regarding utilization of RFT to further enhance the reasoning capabilities of MLLMs. Excitingly, the community has already achieved remarkable successes on this topic, including diverse modalities, diverse tasks and domains, better training algorithms, abundant benchmarks and thriving engineering frameworks. Following discussions with the MLLM, LLM, and RL communities, we believe that the following five points still warrant further research:

\textbf{\color{C1}\uline{TO DO 1:  Achieve Better Generalization across Modalities, Tasks and Domains}. }Although considerable research has focused on cross-task reasoning, existing efforts remain limited to specific domains and modalities; furthermore, the scope of these tasks is limited, typically encompassing only two or three tasks. However, in the pursuit of AGI, we have always aspired to develop a single model capable of adapting to a diverse array of modalities, tasks, and domains. Therefore, research on generalizable reasoning holds significant value. X-Reasoner \citep{liu2025x} is a pioneer in this area, demonstrating that general-domain text-based post-training can enable generalizable reasoning and that performance in specialized domains could be further enhanced through training on domain-specific (e.g., medical-specific) text-only data. Moreover, it can be observed that there are still more points worth exploring in this area. Firstly, {\color{C1}modalities other than textual and visual} have not been addressed; therefore, future work could further explore generalizable reasoning capabilities for more complex modalities, such as auditory and omni-multimodal. Furthermore, the {\color{C1}reasoning capability generalization for broader tasks}, such as progressing from the perceptual vision task (image) to the temporal vision task (video), deserves further exploration within the community. Lastly, {\color{C1}generalization of the reasoning capability across broader domains}, exemplified by the shift from general-domain to embodied-specific settings, remains an underexplored area that requires further systematic investigation.  

\textbf{\color{C1}\uline{TO DO 2: Combine the Outcome Reward Paradigm and the Process Reward Paradigm}. }The outcome reward paradigm offers high efficiency and ease of implementation; however, the sparsity of its rewards provides no intermediate feedback during the reasoning process. For the process reward paradigm, while dense rewards are available for intermediate reasoning steps, training of the process reward model (PRM) remains relatively challenging and unstable. Therefore, the community could consider integrating the outcome reward paradigm with the process reward paradigm. On one hand, {\color{C1}PRM training could be powered by the outcome reward paradigm}. Regarding multimodal reward model training, R1-Reward \citep{zhang2025r1reward}, UnifiedReward-Think \citep{wang2025unified} and Skywork-VL Reward \citep{wang2025skywork} have conducted pioneering research, demonstrating that RFT could lead to more stable training dynamics and enhanced performance; therefore, future research could investigate the integration of outcome reward paradigm to enhance PRM training. On the other hand, {\color{C1}further exploration is warranted regarding the provision of effective and dense rewards in the outcome reward paradigm.} StepGRPO \citep{zhang2025r1} represents a pioneering approach to this area, notably by incorporating dense step-wise rewards; however, it is limited to the vision mathematical reasoning task, and the applicability of such a methodology to other tasks, domains, and modalities requires further investigation.

\textbf{\color{C1}\uline{TO DO 3: Pay More Attention towards the Safety of Reasoning MLLMs}. }Safeguarding MLLMs against security vulnerabilities and adversarial threats is a critical research area that has been widely explored in the community \citep{pi2024mllm, gou2024eyes, gu2024mllmguard, liu2024safety}. Recently, it has been indicated that reasoning LLMs present novel safety challenges stemming from their training algorithms, exposure to adversarial attacks during inference, and vulnerabilities inherent in their deployment environments \citep{zhou2025hidden, jiang2025safechain,yao2025mousetrap}. Nevertheless, research specifically on safety for reasoning MLLMs remains notably limited, which is a critical area that demands increased attention from the community. Future research could further focus on developing {\color{C1}advanced detection and defense mechanisms} specifically designed for reasoning MLLMs. According to \citep{zhang2025100}, this point can generally be divided into three components. Firstly, {\color{C1}reward hacking}, a persistent challenge within the community \citep{amodei2016concrete,weng2024rewardhack}, warrants further attention and effort. Moreover, the exploration of {\color{C1}jailbreak attacks and defenses in reasoning MLLMs} deserves greater focus within the community. Lastly, {\color{C1}the issue of  overthinking}, as highlighted by pioneering works such as No-Thinking-RL \citep{li2025think} and FAST \citep{xiao2025fast}, is also a critical challenge within the community that could be further investigated across more diverse modalities, tasks, and domains.

\textbf{\color{C1}\uline{TO DO 4: Investigate More Data Augmentation Attempts for Multimodality}. }Data augmentation (DA) has been demonstrated to be an effective technique for MLLMs' training \citep{sun2023aligning,zhu2024self} and could potentially enhance the performance and robustness of models. In RFT settings for MLLMs, data is often scarce; therefore, internal data augmentation is likely to enhance the model's perception capabilities. NoisyRollout \citep{liu2025noisyrollout} pioneers in this area, which demonstrates that incorporating Gaussian noise during training could enhance the reasoning performance on visual mathematical task. Therefore, further exploration in the following points might be valuable. Firstly, {\color{C1}appropriate DA methods for a broader range of visual tasks} (such as the visual counting task) could be further explored. Moreover, it is also worthwhile to further explore {\color{C1}more appropriate and diverse DA methods} (such as RandomResizedCrop, RandomCrop, CenterCrop, RandFlip, RandomAffine, RandomInvert, and so on \citep{zhu2024self}) for all these tasks. Lastly, the potential for {\color{C1}applying DA methods to other modalities} and evaluating their effectiveness in these contexts merits further investigation.

\textbf{\color{C1}\uline{TO DO 5: Explore Better Algorithms, Reward Paradigms, and Beyond}. }As previously discussed, the community has made substantial progress in {\color{C1}developing improved training algorithms}. Furthermore, this should continue to be one of the key areas of focus for community efforts. Regarding reward paradigms, rule-based rewards are typically employed in current algorithms. In future research, it is valuable to further {\color{C1}explore automatic frameworks for designing task-specific reward functions}. Finally, exploring the implementation of reinforcement fine-tuned reasoning MLLMs {\color{C1}across diverse academic disciplines} (such as architecture, aerospace, electric engineering, and so on) is a promising field that necessitates collaborative efforts from various disciplinary communities. 

\begin{ack}
Haoyuan Sun extends his sincere gratitude to all community members who provided valuable supplementary support to the project. This work was supported by the Natural Science Foundation of Shenzhen (No.JCYJ20230807111604008, No.JCYJ20240813112007010), the Natural Science Foundation of Guangdong Province (No.2024A1515010003), National Key Research and Development Program of China (No.2022YFB4701400) and Cross-disciplinary Fund for Research and Innovation (No.JC2024002) of Tsinghua SIGS.
\end{ack}

\bibliographystyle{unsrtnat}  


\newpage
\appendix

\section{Summary of works done on RFT for MLLMs}
\label{Summary of works done on RFT for MLLMs}
\subsection{Vision (Image)}
\textbf{Papers}

[2505] [OpenThinkIMG \citep{su2025openthinkimg} ]  
\href{https://arxiv.org/abs/2505.08617}{OpenThinkIMG: Learning to Think with Images via Visual Tool Reinforcement Learning} [\href{https://huggingface.co/Warrieryes/OpenThinkIMG-Chart-Qwen2-2B-VL}{\raisebox{-0.35ex}{\protect\includegraphics[height=1em]{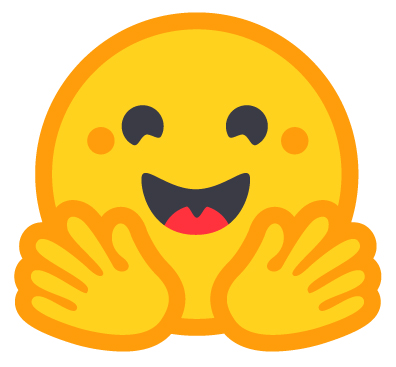}}Model}] [\href{https://huggingface.co/collections/Warrieryes/openthinkimg-68244a63e97a24d9b7ffcde9}{\raisebox{-0.35ex}{\protect\includegraphics[height=1em]{huggingface.jpg}}Datasets}] [\href{https://github.com/zhaochen0110/OpenThinkIMG}{\raisebox{-0.35ex}{\protect\includegraphics[height=1em]{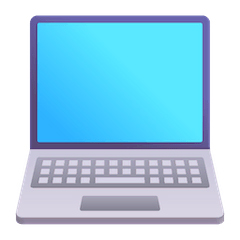}}Code}]

[2505] [DanceGRPO (Gen) \citep{xue2025dancegrpo} ] \href{https://arxiv.org/abs/2505.07818}{DanceGRPO: Unleashing GRPO on Visual Generation} [\href{https://dancegrpo.github.io/}{\raisebox{-0.35ex}{\protect\includegraphics[height=1em]{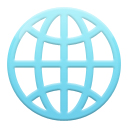}}Project}]
[\href{https://github.com/XueZeyue/DanceGRPO}{\raisebox{-0.35ex}{\protect\includegraphics[height=1em]{laptop.jpg}}Code}]

[2505] [Flow-GRPO (Gen) \citep{liu2025flow} ] \href{https://arxiv.org/abs/2505.05470}{Flow-GRPO: Training Flow Matching Models via Online RL} 
[\href{https://huggingface.co/jieliu}{\raisebox{-0.35ex}{\protect\includegraphics[height=1em]{huggingface.jpg}} Models}] [\href{https://github.com/yifan123/flow_grpo}{\raisebox{-0.35ex}{\protect\includegraphics[height=1em]{laptop.jpg}}Code}]

[2505] [X-Reasoner \citep{liu2025x} ] \href{https://arxiv.org/abs/2505.03981}{X-Reasoner: Towards Generalizable Reasoning Across Modalities and Domains} [\href{https://github.com/microsoft/x-reasoner}{\raisebox{-0.35ex}{\protect\includegraphics[height=1em]{laptop.jpg}}Code}]

[2505] [T2I-R1 (Gen) \citep{jiang2025t2i} ] \href{https://arxiv.org/abs/2505.00703}{T2I-R1: Reinforcing Image Generation with Collaborative Semantic-level and Token-level CoT} [\href{https://github.com/CaraJ7/T2I-R1}{\raisebox{-0.35ex}{\protect\includegraphics[height=1em]{laptop.jpg}}Code}]

[2504] [FAST \citep{xiao2025fast} ] \href{https://arxiv.org/abs/2504.18458}{Fast-Slow Thinking for Large Vision-Language Model Reasoning} [\href{https://github.com/Mr-Loevan/FAST}{\raisebox{-0.35ex}{\protect\includegraphics[height=1em]{laptop.jpg}}Code}]

[2504] [Skywork R1V2 \citep{wei2025skywork} ] \href{https://arxiv.org/abs/2504.16656}{Skywork R1V2: Multimodal Hybrid Reinforcement Learning for Reasoning}
[\href{https://huggingface.co/collections/Skywork/skywork-r1v2-68075a3d947a5ae160272671}{\raisebox{-0.35ex}{\protect\includegraphics[height=1em]{huggingface.jpg}}Models}] 
[\href{https://github.com/SkyworkAI/Skywork-R1V}{\raisebox{-0.35ex}{\protect\includegraphics[height=1em]{laptop.jpg}}Code}]

[2504] [Relation-R1 \citep{li2025relation} ]
\href{https://arxiv.org/abs/2504.14642}{Relation-R1: Cognitive Chain-of-Thought Guided Reinforcement Learning for Unified Relational Comprehension}
[\href{https://github.com/HKUST-LongGroup/Relation-R1}{\raisebox{-0.35ex}{\protect\includegraphics[height=1em]{laptop.jpg}}Code}]

[2504] [R1-SGG \citep{chen2025compile} ] \href{https://arxiv.org/abs/2504.13617}{Compile Scene Graphs with Reinforcement Learning}
[\href{https://github.com/gpt4vision/R1-SGG}{\raisebox{-0.35ex}{\protect\includegraphics[height=1em]{laptop.jpg}}Code}]

[2504] [NoisyRollout \citep{liu2025noisyrollout} ] \href{https://arxiv.org/abs/2504.13055}{Reinforcing Visual Reasoning with Data Augmentation}
[\href{https://huggingface.co/collections/xyliu6/noisyrollout-67ff992d1cf251087fe021a2}{\raisebox{-0.35ex}{\protect\includegraphics[height=1em]{huggingface.jpg}}Models}] [\href{https://huggingface.co/collections/xyliu6/noisyrollout-67ff992d1cf251087fe021a2}{\raisebox{-0.35ex}{\protect\includegraphics[height=1em]{huggingface.jpg}}Datasets}]
[\href{https://github.com/John-AI-Lab/NoisyRollout}{\raisebox{-0.35ex}{\protect\includegraphics[height=1em]{laptop.jpg}}Code}]

[2504] [SimpleAR (Gen) \citep{wang2025simplear} ] \href{https://arxiv.org/abs/2504.11455}{SimpleAR: Pushing the Frontier of Autoregressive Visual Generation through Pretraining, SFT, and RL}
[\href{https://huggingface.co/collections/Daniel0724/simplear-6805053f5b4b9961ac025136}{\raisebox{-0.35ex}{\protect\includegraphics[height=1em]{huggingface.jpg}}Models}] 
[\href{https://github.com/wdrink/SimpleAR}{\raisebox{-0.35ex}{\protect\includegraphics[height=1em]{laptop.jpg}}Code}]

[2504] [VL-Rethinker \citep{wang2025vl} ] \href{https://arxiv.org/abs/2504.08837}{VL-Rethinker: Incentivizing Self-Reflection of Vision-Language Models with Reinforcement Learning} [\href{https://tiger-ai-lab.github.io/VL-Rethinker}{\raisebox{-0.35ex}{\protect\includegraphics[height=1em]{earth.jpg}}Project}] 
[\href{https://huggingface.co/collections/TIGER-Lab/vl-rethinker-67fdc54de07c90e9c6c69d09}{\raisebox{-0.35ex}{\protect\includegraphics[height=1em]{huggingface.jpg}}Models}] [\href{https://huggingface.co/datasets/TIGER-Lab/ViRL39K}{\raisebox{-0.35ex}{\protect\includegraphics[height=1em]{huggingface.jpg}}Dataset}]
[\href{https://github.com/TIGER-AI-Lab/VL-Rethinker}{\raisebox{-0.35ex}{\protect\includegraphics[height=1em]{laptop.jpg}}Code}]

[2504] [Kimi-VL \citep{team2025kimi-vl} ] \href{https://arxiv.org/abs/2504.07491}{Kimi-VL Technical Report} [\href{https://github.com/MoonshotAI/Kimi-VL}{\raisebox{-0.35ex}{\protect\includegraphics[height=1em]{earth.jpg}}Project}] 
[\href{https://huggingface.co/collections/moonshotai/kimi-vl-a3b-67f67b6ac91d3b03d382dd85}{\raisebox{-0.35ex}{\protect\includegraphics[height=1em]{huggingface.jpg}}Models}] [\href{https://huggingface.co/spaces/moonshotai/Kimi-VL-A3B-Thinking}{\raisebox{-0.35ex}{\protect\includegraphics[height=1em]{huggingface.jpg}}Demo}]
[\href{https://github.com/MoonshotAI/Kimi-VL}{\raisebox{-0.35ex}{\protect\includegraphics[height=1em]{laptop.jpg}}Code}]

[2504] [VLAA-Thinking \citep{chen2025sft} ] \href{https://arxiv.org/abs/2504.11468}{SFT or RL? An Early Investigation into Training R1-Like Reasoning Large Vision-Language Models} 
[\href{https://huggingface.co/collections/UCSC-VLAA/vlaa-thinker-67eda033419273423d77249e}{\raisebox{-0.35ex}{\protect\includegraphics[height=1em]{huggingface.jpg}}Models}] [\href{https://huggingface.co/datasets/UCSC-VLAA/VLAA-Thinking}{\raisebox{-0.35ex}{\protect\includegraphics[height=1em]{huggingface.jpg}}Dataset}]
[\href{https://github.com/UCSC-VLAA/VLAA-Thinking}{\raisebox{-0.35ex}{\protect\includegraphics[height=1em]{laptop.jpg}}Code}]

[2504] [Perception-R1 \citep{yu2025perception} ] \href{https://arxiv.org/abs/2504.07954}{Perception-R1: Pioneering Perception Policy with Reinforcement Learning} 
[\href{https://huggingface.co/collections/Kangheng/perception-r1-67f6b14f89d307a0ece985af}{\raisebox{-0.35ex}{\protect\includegraphics[height=1em]{huggingface.jpg}}Models}] [\href{https://huggingface.co/collections/Kangheng/perception-r1-67f6b14f89d307a0ece985af}{\raisebox{-0.35ex}{\protect\includegraphics[height=1em]{huggingface.jpg}}Datasets}]
[\href{https://github.com/linkangheng/PR1}{\raisebox{-0.35ex}{\protect\includegraphics[height=1em]{laptop.jpg}}Code}]

[2504] [SoTA with Less \citep{wang2025sota} ] \href{https://arxiv.org/abs/2504.07934}{SoTA with Less: MCTS-Guided Sample Selection for Data-Efficient Visual Reasoning Self-Improvement} 
[\href{https://huggingface.co/russwang/ThinkLite-VL-7B}{\raisebox{-0.35ex}{\protect\includegraphics[height=1em]{huggingface.jpg}}Model}] [\href{https://huggingface.co/collections/russwang/thinklite-vl-67f88c6493f8a7601e73fe5a}{\raisebox{-0.35ex}{\protect\includegraphics[height=1em]{huggingface.jpg}}Datasets}]
[\href{https://github.com/si0wang/ThinkLite-VL}{\raisebox{-0.35ex}{\protect\includegraphics[height=1em]{laptop.jpg}}Code}]

[2504] [VLM-R1 \citep{shen2025vlm} ] \href{https://arxiv.org/abs/2504.07615}{VLM-R1: A Stable and Generalizable R1-style Large Vision-Language Model} 
[\href{https://huggingface.co/omlab/Qwen2.5VL-3B-VLM-R1-REC-500steps}{\raisebox{-0.35ex}{\protect\includegraphics[height=1em]{huggingface.jpg}}Model}] [\href{https://huggingface.co/datasets/omlab/VLM-R1}{\raisebox{-0.35ex}{\protect\includegraphics[height=1em]{huggingface.jpg}}Dataset}]
[\href{https://huggingface.co/spaces/omlab/VLM-R1-Referral-Expression}{\raisebox{-0.35ex}{\protect\includegraphics[height=1em]{huggingface.jpg}}Demo}]
[\href{https://github.com/om-ai-lab/VLM-R1}{\raisebox{-0.35ex}{\protect\includegraphics[height=1em]{laptop.jpg}}Code}]

[2504] [CrowdVLM-R1 \citep{wang2025crowdvlm} ] \href{https://arxiv.org/abs/2504.03724}{CrowdVLM-R1: Expanding R1 Ability to Vision Language Model for Crowd Counting using Fuzzy Group Relative Policy Reward} 
[\href{https://huggingface.co/datasets/yeyimilk/CrowdVLM-R1-data}{\raisebox{-0.35ex}{\protect\includegraphics[height=1em]{huggingface.jpg}}Dataset}]
[\href{https://github.com/yeyimilk/CrowdVLM-R1}{\raisebox{-0.35ex}{\protect\includegraphics[height=1em]{laptop.jpg}}Code}]

[2504] [MAYE \citep{ma2025maye} ] \href{https://www.arxiv.org/abs/2504.02587}{Rethinking RL Scaling for Vision Language Models: A Transparent, From-Scratch Framework and Comprehensive Evaluation Scheme} 
[\href{https://huggingface.co/datasets/ManTle/MAYE}{\raisebox{-0.35ex}{\protect\includegraphics[height=1em]{huggingface.jpg}}Dataset}]
[\href{https://github.com/GAIR-NLP/MAYE}{\raisebox{-0.35ex}{\protect\includegraphics[height=1em]{laptop.jpg}}Code}]

[2503] [Q-Insight \citep{li2025q} ] \href{https://arxiv.org/abs/2503.22679}{Q-Insight: Understanding Image Quality via Visual Reinforcement Learning} 
[\href{https://github.com/lwq20020127/Q-Insight}{\raisebox{-0.35ex}{\protect\includegraphics[height=1em]{laptop.jpg}}Code}]

[2503] [Reason-RFT \citep{tan2025reason} ] \href{https://arxiv.org/abs/2503.20752}{Reason-RFT: Reinforcement Fine-Tuning for Visual Reasoning} 
[\href{https://tanhuajie.github.io/ReasonRFT}{\raisebox{-0.35ex}{\protect\includegraphics[height=1em]{earth.jpg}}Project}] 
[\href{https://huggingface.co/datasets/tanhuajie2001/Reason-RFT-CoT-Dataset}{\raisebox{-0.35ex}{\protect\includegraphics[height=1em]{huggingface.jpg}}Dataset}]
[\href{https://github.com/tanhuajie/Reason-RFT}{\raisebox{-0.35ex}{\protect\includegraphics[height=1em]{laptop.jpg}}Code}]

[2503] [OpenVLThinker \citep{deng2025openvlthinker} ] \href{https://arxiv.org/abs/2503.17352}{OpenVLThinker: An Early Exploration to Vision-Language Reasoning via Iterative Self-Improvement} 
[\href{https://huggingface.co/ydeng9/OpenVLThinker-7B}{\raisebox{-0.35ex}{\protect\includegraphics[height=1em]{huggingface.jpg}}Model}]
[\href{https://github.com/yihedeng9/OpenVLThinker}{\raisebox{-0.35ex}{\protect\includegraphics[height=1em]{laptop.jpg}}Code}]

[2503] [Think or Not Think \citep{li2025think} ] \href{https://arxiv.org/abs/2503.16188}{Think or Not Think: A Study of Explicit Thinking in Rule-Based Visual Reinforcement Fine-Tuning} 
[\href{https://huggingface.co/afdsafas}{\raisebox{-0.35ex}{\protect\includegraphics[height=1em]{huggingface.jpg}}Models}]
[\href{https://huggingface.co/afdsafas}{\raisebox{-0.35ex}{\protect\includegraphics[height=1em]{huggingface.jpg}}Datasets}]
[\href{https://github.com/minglllli/CLS-RL}{\raisebox{-0.35ex}{\protect\includegraphics[height=1em]{laptop.jpg}}Code}]

[2503] [OThink-MR1 \citep{liu2025othink} ] \href{https://arxiv.org/abs/2503.16081}{OThink-MR1: Stimulating multimodal generalized reasoning capabilities via dynamic reinforcement learning} 

[2503] [R1-VL \citep{zhang2025r1} ] \href{https://arxiv.org/abs/2503.12937}{R1-VL: Learning to Reason with Multimodal Large Language Models via Step-wise Group Relative Policy Optimization} 
[\href{https://huggingface.co/jingyiZ00}{\raisebox{-0.35ex}{\protect\includegraphics[height=1em]{huggingface.jpg}}Model}]
[\href{https://github.com/jingyi0000/R1-VL}{\raisebox{-0.35ex}{\protect\includegraphics[height=1em]{laptop.jpg}}Code}]

[2503] [Skywork R1V \citep{wei2025skywork} ] \href{https://github.com/SkyworkAI/Skywork-R1V/blob/main/Skywork_R1V.pdf}{Skywork R1V: Pioneering Multimodal Reasoning with Chain-of-Thought} 
[\href{https://huggingface.co/Skywork/Skywork-R1V-38B}{\raisebox{-0.35ex}{\protect\includegraphics[height=1em]{huggingface.jpg}}Model}]
[\href{https://github.com/SkyworkAI/Skywork-R1V}{\raisebox{-0.35ex}{\protect\includegraphics[height=1em]{laptop.jpg}}Code}]

[2503] [R1-Onevision \citep{yang2025r1} ] \href{https://arxiv.org/abs/2503.10615}{R1-Onevision: Advancing Generalized Multimodal Reasoning through Cross-Modal Formalization} 
[\href{https://huggingface.co/Fancy-MLLM/R1-Onevision-7B}{\raisebox{-0.35ex}{\protect\includegraphics[height=1em]{huggingface.jpg}}Model}] [\href{https://huggingface.co/datasets/Fancy-MLLM/R1-Onevision}{\raisebox{-0.35ex}{\protect\includegraphics[height=1em]{huggingface.jpg}}Dataset}]
[\href{https://huggingface.co/spaces/Fancy-MLLM/R1-Onevision}{\raisebox{-0.35ex}{\protect\includegraphics[height=1em]{huggingface.jpg}}Demo}]
[\href{https://github.com/Fancy-MLLM/R1-Onevision}{\raisebox{-0.35ex}{\protect\includegraphics[height=1em]{laptop.jpg}}Code}]

[2503] [VisualPRM \citep{wang2025visualprm} ] \href{https://arxiv.org/abs/2503.10291v1}{VisualPRM: An Effective Process Reward Model for Multimodal Reasoning} 
[\href{https://internvl.github.io/blog/2025-03-13-VisualPRM}{\raisebox{-0.35ex}{\protect\includegraphics[height=1em]{earth.jpg}}Project}] 
[\href{https://huggingface.co/OpenGVLab/VisualPRM-8B}{\raisebox{-0.35ex}{\protect\includegraphics[height=1em]{huggingface.jpg}}Model}] [\href{https://huggingface.co/datasets/OpenGVLab/VisualPRM400K}{\raisebox{-0.35ex}{\protect\includegraphics[height=1em]{huggingface.jpg}}Dataset}]
[\href{https://huggingface.co/datasets/OpenGVLab/VisualProcessBench}{\raisebox{-0.35ex}{\protect\includegraphics[height=1em]{huggingface.jpg}}Benchmark}]

[2503] [LMM-R1 \citep{peng2025lmm} ] \href{https://arxiv.org/abs/2503.07536}{LMM-R1: Empowering 3B LMMs with Strong Reasoning Abilities Through Two-Stage Rule-Based RL} [\href{https://github.com/TideDra/lmm-r1}{\raisebox{-0.35ex}{\protect\includegraphics[height=1em]{laptop.jpg}}Code}]

[2503] [Curr-ReFT \citep{deng2025boosting} ] \href{https://arxiv.org/abs/2503.07065}{Boosting the Generalization and Reasoning of Vision Language Models with Curriculum Reinforcement Learning} 
[\href{https://huggingface.co/ZTE-AIM}{\raisebox{-0.35ex}{\protect\includegraphics[height=1em]{huggingface.jpg}}Models}] [\href{https://huggingface.co/datasets/ZTE-AIM/Curr-ReFT-data}{\raisebox{-0.35ex}{\protect\includegraphics[height=1em]{huggingface.jpg}}Dataset}]
[\href{https://github.com/ding523/Curr_REFT}{\raisebox{-0.35ex}{\protect\includegraphics[height=1em]{laptop.jpg}}Code}]

[2503] [VisualThinker-R1-Zero \citep{zhou2025r1} ] \href{https://arxiv.org/abs/2503.05132}{R1-Zero's "Aha Moment" in Visual Reasoning on a 2B Non-SFT Model} [\href{https://github.com/turningpoint-ai/VisualThinker-R1-Zero}{\raisebox{-0.35ex}{\protect\includegraphics[height=1em]{laptop.jpg}}Code}]

[2503] [Vision-R1 \citep{huang2025vision} ] \href{https://arxiv.org/abs/2503.06749}{Vision-R1: Incentivizing Reasoning Capability in Multimodal Large Language Models} [\href{https://github.com/Osilly/Vision-R1}{\raisebox{-0.35ex}{\protect\includegraphics[height=1em]{laptop.jpg}}Code}]

[2503] [Seg-Zero \citep{liu2025seg} ] \href{https://arxiv.org/abs/2503.06520}{Seg-Zero: Reasoning-Chain Guided Segmentation via Cognitive Reinforcement} 
[\href{https://huggingface.co/Ricky06662/Seg-Zero-7B}{\raisebox{-0.35ex}{\protect\includegraphics[height=1em]{huggingface.jpg}}Model}] [\href{https://huggingface.co/datasets/Ricky06662/refCOCOg_2k_840}{\raisebox{-0.35ex}{\protect\includegraphics[height=1em]{huggingface.jpg}}Dataset}]
[\href{https://github.com/dvlab-research/Seg-Zero}{\raisebox{-0.35ex}{\protect\includegraphics[height=1em]{laptop.jpg}}Code}]

[2503] [MM-Eureka \citep{meng2025mm} ] \href{https://github.com/ModalMinds/MM-EUREKA/blob/main/MM_Eureka_paper.pdf}{MM-Eureka: Exploring Visual Aha Moment with Rule-based Large-scale Reinforcement Learning} 
[\href{https://huggingface.co/FanqingM}{\raisebox{-0.35ex}{\protect\includegraphics[height=1em]{huggingface.jpg}}Models}] [\href{https://huggingface.co/datasets/FanqingM/MM-Eureka-Dataset}{\raisebox{-0.35ex}{\protect\includegraphics[height=1em]{huggingface.jpg}}Dataset}]
[\href{https://github.com/ModalMinds/MM-EUREKA}{\raisebox{-0.35ex}{\protect\includegraphics[height=1em]{laptop.jpg}}Code}]

[2503] [Visual-RFT \citep{liu2025visual} ] \href{https://arxiv.org/abs/2503.01785}{Visual-RFT: Visual Reinforcement Fine-Tuning} 
[\href{https://github.com/Liuziyu77/Visual-RFT}{\raisebox{-0.35ex}{\protect\includegraphics[height=1em]{earth.jpg}}Project}] 
[\href{https://huggingface.co/collections/laolao77/virft-datasets-67bc271b6f2833eccc0651df}{\raisebox{-0.35ex}{\protect\includegraphics[height=1em]{huggingface.jpg}}Data sets}][\href{https://github.com/Liuziyu77/Visual-RFT}{\raisebox{-0.35ex}{\protect\includegraphics[height=1em]{laptop.jpg}}Code}]

[2501] [PARM++ (Gen) \citep{guo2025can} ] \href{https://arxiv.org/abs/2501.13926}{Can We Generate Images with CoT? Let’s Verify and Reinforce Image Generation Step by Step}
[\href{https://github.com/ZiyuGuo99/Image-Generation-CoT}{\raisebox{-0.35ex}{\protect\includegraphics[height=1em]{earth.jpg}}Project}] [\href{https://huggingface.co/ZiyuG/Image-Generation-CoT}{\raisebox{-0.35ex}{\protect\includegraphics[height=1em]{huggingface.jpg}}Model}] 
[\href{https://github.com/ZiyuGuo99/Image-Generation-CoT}{\raisebox{-0.35ex}{\protect\includegraphics[height=1em]{laptop.jpg}}Code}]

[2501] [Kimi k1.5 \citep{team2025kimi} ] \href{https://arxiv.org/abs/2501.12599}{Kimi k1.5: Scaling Reinforcement Learning with LLMs} 
[\href{https://github.com/MoonshotAI/Kimi-k1.5}{\raisebox{-0.35ex}{\protect\includegraphics[height=1em]{earth.jpg}}Project}] 

[2501] [Virgo \citep{du2025virgo} ] \href{https://arxiv.org/abs/2501.01904v2}{Virgo: A Preliminary Exploration on Reproducing o1-like MLLM} 
[\href{https://huggingface.co/RUC-AIBOX/Virgo-72B}{\raisebox{-0.35ex}{\protect\includegraphics[height=1em]{huggingface.jpg}}Model}] 
[\href{https://github.com/RUCAIBox/Virgo}{\raisebox{-0.35ex}{\protect\includegraphics[height=1em]{laptop.jpg}}Code}]

[2412] [Mulberry \citep{yao2024mulberry} ] \href{https://arxiv.org/abs/2412.18319}{Mulberry: Empowering MLLM with o1-like Reasoning and Reflection via Collective Monte Carlo Tree Search} 
[\href{https://huggingface.co/HuanjinYao/Mulberry_llava_8b}{\raisebox{-0.35ex}{\protect\includegraphics[height=1em]{huggingface.jpg}}Model}] 
[\href{https://github.com/HJYao00/Mulberry}{\raisebox{-0.35ex}{\protect\includegraphics[height=1em]{laptop.jpg}}Code}]

[2411] [Insight-V \citep{dong2024insight} ] \href{https://arxiv.org/abs/2411.14432}{Insight-V: Exploring Long-Chain Visual Reasoning with Multimodal Large Language Models}
[\href{https://huggingface.co/collections/THUdyh/insight-v-673f5e1dd8ab5f2d8d332035}{\raisebox{-0.35ex}{\protect\includegraphics[height=1em]{huggingface.jpg}}Model}] 
[\href{https://github.com/dongyh20/Insight-V}{\raisebox{-0.35ex}{\protect\includegraphics[height=1em]{laptop.jpg}}Code}]

[2411] [InternVL2-MPO \citep{wang2024enhancing} ] \href{https://arxiv.org/abs/2411.10442}{Enhancing the Reasoning Ability of Multimodal Large Language Models via Mixed Preference Optimization}
[\href{https://internvl.github.io/blog/2024-11-14-InternVL-2.0-MPO/}{\raisebox{-0.35ex}{\protect\includegraphics[height=1em]{earth.jpg}}Project}] [\href{https://huggingface.co/OpenGVLab/InternVL2-8B-MPO}{\raisebox{-0.35ex}{\protect\includegraphics[height=1em]{huggingface.jpg}}Model}] 
[\href{https://github.com/OpenGVLab/InternVL/tree/main/internvl_chat/shell/internvl2.0_mpo}{\raisebox{-0.35ex}{\protect\includegraphics[height=1em]{laptop.jpg}}Code}]

\textbf{Open-Source Projects (Repository without Paper)}

[R1-V \citep{chen2025r1v} ] 
[\href{https://github.com/Deep-Agent/R1-V}{\raisebox{-0.35ex}{\protect\includegraphics[height=1em]{laptop.jpg}}Code}] [\href{https://huggingface.co/collections/MMInstruction/r1-v-67aae24fa56af9d2e2755f82}{\raisebox{-0.35ex}{\protect\includegraphics[height=1em]{huggingface.jpg}}Datasets}] [\href{https://deepagent.notion.site/rlvr-in-vlms}{\raisebox{-0.35ex}{\protect\includegraphics[height=1em]{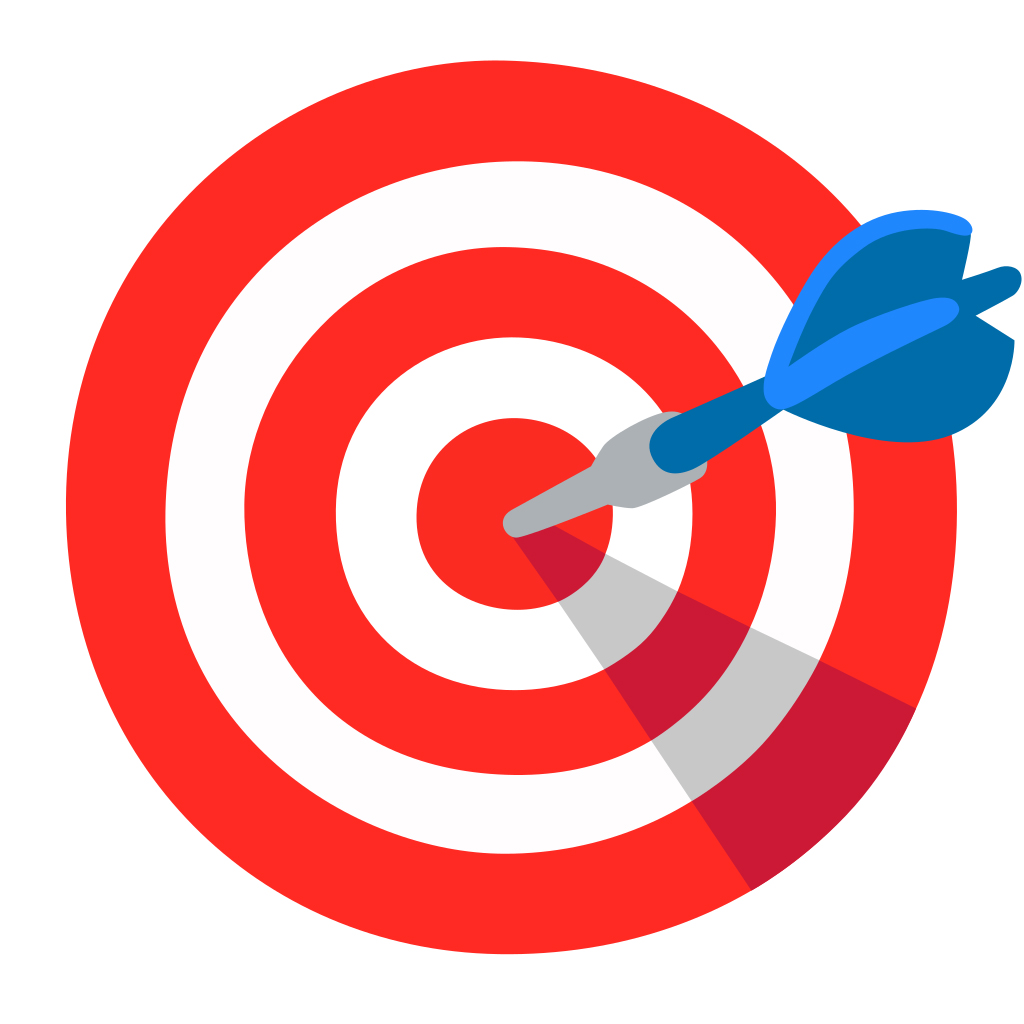}}Blog}] 

[Multimodal Open R1 \citep{openr1} ] [\href{https://github.com/EvolvingLMMs-Lab/open-r1-multimodal}{\raisebox{-0.35ex}{\protect\includegraphics[height=1em]{laptop.jpg}}Code}] [\href{https://huggingface.co/lmms-lab/Qwen2-VL-2B-GRPO-8k}{\raisebox{-0.35ex}{\protect\includegraphics[height=1em]{huggingface.jpg}}Model}] [\href{https://huggingface.co/datasets/lmms-lab/multimodal-open-r1-8k-verified}{\raisebox{-0.35ex}{\protect\includegraphics[height=1em]{huggingface.jpg}}Dataset}] 

[MMR1 \citep{MMR1-Math2025} ] [\href{https://github.com/LengSicong/MMR1}{\raisebox{-0.35ex}{\protect\includegraphics[height=1em]{laptop.jpg}}Code}] [\href{https://huggingface.co/MMR1/MMR1-Math-v0-7B}{\raisebox{-0.35ex}{\protect\includegraphics[height=1em]{huggingface.jpg}}Model}] [\href{https://huggingface.co/datasets/MMR1/MMR1-Math-RL-Data-v0}{\raisebox{-0.35ex}{\protect\includegraphics[height=1em]{huggingface.jpg}}Dataset}]  

[R1-Multimodal-Journey \citep{meng2025mm} ] [\href{https://github.com/FanqingM/R1-Multimodal-Journey}{\raisebox{-0.35ex}{\protect\includegraphics[height=1em]{laptop.jpg}}Code}] 

[R1-Vision \citep{yu25r1vision} ] [\href{https://github.com/yuyq96/R1-Vision}{\raisebox{-0.35ex}{\protect\includegraphics[height=1em]{laptop.jpg}}Code}] [\href{https://huggingface.co/collections/yuyq96/r1-vision-67a6fb7898423dca453efa83}{\raisebox{-0.35ex}{\protect\includegraphics[height=1em]{huggingface.jpg}}Cold-Start Datasets}]

[Ocean-R1 \citep{ming2025oceanr1} ] [\href{https://github.com/VLM-RL/Ocean-R1}{\raisebox{-0.35ex}{\protect\includegraphics[height=1em]{laptop.jpg}}Code}] [\href{https://huggingface.co/minglingfeng}{\raisebox{-0.35ex}{\protect\includegraphics[height=1em]{huggingface.jpg}}Models}] [\href{https://huggingface.co/minglingfeng}{\raisebox{-0.35ex}{\protect\includegraphics[height=1em]{huggingface.jpg}}Datasets}] 

[R1V-Free \citep{Cheng_R1V-Free_Advancing_Open-World_2025} ] [\href{https://github.com/VLM-RL/Ocean-R1}{\raisebox{-0.35ex}{\protect\includegraphics[height=1em]{laptop.jpg}}Code}] [\href{https://huggingface.co/collections/Exgc/r1v-free-67f769feedffab8761b8f053}{\raisebox{-0.35ex}{\protect\includegraphics[height=1em]{huggingface.jpg}}Models}] [\href{https://huggingface.co/datasets/Exgc/R1V-Free_RLHFV}{\raisebox{-0.35ex}{\protect\includegraphics[height=1em]{huggingface.jpg}}Dataset}] 

[SeekWorld \citep{seekworld2025} ] [\href{https://github.com/TheEighthDay/SeekWorld}{\raisebox{-0.35ex}{\protect\includegraphics[height=1em]{laptop.jpg}}Code}] [\href{https://huggingface.co/TheEighthDay/SeekWorld_RL_PLUS}{\raisebox{-0.35ex}{\protect\includegraphics[height=1em]{huggingface.jpg}}Model}] [\href{https://huggingface.co/datasets/TheEighthDay/SeekWorld}{\raisebox{-0.35ex}{\protect\includegraphics[height=1em]{huggingface.jpg}}Dataset}] [\href{https://huggingface.co/spaces/TheEighthDay/SeekWorld_APP}{\raisebox{-0.35ex}{\protect\includegraphics[height=1em]{huggingface.jpg}}Demo}] 

[R1-Track \citep{wang2025r1track} ] [\href{https://github.com/Wangbiao2/R1-Track}{\raisebox{-0.35ex}{\protect\includegraphics[height=1em]{laptop.jpg}}Code}] [\href{https://huggingface.co/WangBiao}{\raisebox{-0.35ex}{\protect\includegraphics[height=1em]{huggingface.jpg}}Models}] [\href{https://huggingface.co/WangBiao}{\raisebox{-0.35ex}{\protect\includegraphics[height=1em]{huggingface.jpg}}Datasets}]  

\subsection{Vision (Video)}
\textbf{Papers}

[2504] [TinyLLaVA-Video-R1 \citep{zhang2025tinyllava} ]  
\href{https://arxiv.org/abs/2504.09641}{TinyLLaVA-Video-R1: Towards Smaller LMMs for Video Reasoning} [\href{https://huggingface.co/Zhang199/TinyLLaVA-Video-R1}{\raisebox{-0.35ex}{\protect\includegraphics[height=1em]{huggingface.jpg}}Model}] [\href{https://github.com/ZhangXJ199/TinyLLaVA-Video-R1}{\raisebox{-0.35ex}{\protect\includegraphics[height=1em]{laptop.jpg}}Code}]

[2504] [VideoChat-R1 \citep{li2025videochat} ]  
\href{https://arxiv.org/abs/2504.06958}{VideoChat-R1: Enhancing Spatio-Temporal Perception via Reinforcement Fine-Tuning} [\href{https://huggingface.co/collections/OpenGVLab/videochat-r1-67fbe26e4eb08c83aa24643e}{\raisebox{-0.35ex}{\protect\includegraphics[height=1em]{huggingface.jpg}}Model}] [\href{https://github.com/OpenGVLab/VideoChat-R1}{\raisebox{-0.35ex}{\protect\includegraphics[height=1em]{laptop.jpg}}Code}]

[2504] [Spatial-R1 \citep{ouyang2025spatial} ]  
\href{https://arxiv.org/abs/2504.01805}{Spatial-R1: Enhancing MLLMs in Video Spatial Reasoning} [\href{https://huggingface.co/RUBBISHLIKE/SpaceR}{\raisebox{-0.35ex}{\protect\includegraphics[height=1em]{huggingface.jpg}}Model}] [\href{https://huggingface.co/RUBBISHLIKE}{\raisebox{-0.35ex}{\protect\includegraphics[height=1em]{huggingface.jpg}}Datasets}] [\href{https://github.com/OuyangKun10/SpaceR}{\raisebox{-0.35ex}{\protect\includegraphics[height=1em]{laptop.jpg}}Code}]

[2504] [R1-Zero-VSI \citep{liao2025improved} ]  
\href{https://arxiv.org/abs/2504.00883}{Improved Visual-Spatial Reasoning via R1-Zero-Like Training} [\href{https://github.com/zhijie-group/R1-Zero-VSI}{\raisebox{-0.35ex}{\protect\includegraphics[height=1em]{laptop.jpg}}Code}]

[2503] [SEED-Bench-R1 \citep{chen2025exploring} ]  
\href{https://arxiv.org/abs/2503.24376}{Exploring the Effect of Reinforcement Learning on Video Understanding: Insights from SEED-Bench-R1} [\href{https://huggingface.co/datasets/TencentARC/SEED-Bench-R1}{\raisebox{-0.35ex}{\protect\includegraphics[height=1em]{huggingface.jpg}}Dataset}] [\href{https://github.com/TencentARC/SEED-Bench-R1}{\raisebox{-0.35ex}{\protect\includegraphics[height=1em]{laptop.jpg}}Code}]

[2503] [Video-R1 \citep{feng2025video} ]  
\href{https://arxiv.org/abs/2503.21776}{Video-R1: Reinforcing Video Reasoning in MLLMs} [\href{https://huggingface.co/Video-R1/Video-R1-7B}{\raisebox{-0.35ex}{\protect\includegraphics[height=1em]{huggingface.jpg}}Model}] [\href{https://huggingface.co/datasets/Video-R1/Video-R1-data}{\raisebox{-0.35ex}{\protect\includegraphics[height=1em]{huggingface.jpg}}Data set}][\href{https://github.com/tulerfeng/Video-R1}{\raisebox{-0.35ex}{\protect\includegraphics[height=1em]{laptop.jpg}}Code}]

[2503] [TimeZero \citep{wang2025timezero} ]  
\href{https://arxiv.org/abs/2503.13377}{TimeZero: Temporal Video Grounding with Reasoning-Guided LVLM} [\href{https://huggingface.co/wwwyyy/TimeZero-Charades-7B}{\raisebox{-0.35ex}{\protect\includegraphics[height=1em]{huggingface.jpg}} Model}] [\href{https://github.com/www-Ye/TimeZero}{\raisebox{-0.35ex}{\protect\includegraphics[height=1em]{laptop.jpg}}Code}]

\textbf{Open-Source Projects (Repository without Paper)}

[Open R1 Video \citep{wang-2025-open-r1-video} ] 
[\href{https://github.com/Wang-Xiaodong1899/Open-R1-Video}{\raisebox{-0.35ex}{\protect\includegraphics[height=1em]{laptop.jpg}}Code}] [\href{https://huggingface.co/Xiaodong/Open-R1-Video-7B}{\raisebox{-0.35ex}{\protect\includegraphics[height=1em]{huggingface.jpg}}Model}] [\href{https://huggingface.co/datasets/Xiaodong/open-r1-video-4k}{\raisebox{-0.35ex}{\protect\includegraphics[height=1em]{huggingface.jpg}}Dataset}] 

[Temporal-R1 \citep{li2025temporalr1} ] 
[\href{https://github.com/appletea233/Temporal-R1}{\raisebox{-0.35ex}{\protect\includegraphics[height=1em]{laptop.jpg}}Code}] [\href{https://huggingface.co/appletea2333}{\raisebox{-0.35ex}{\protect\includegraphics[height=1em]{huggingface.jpg}}Models}] 

[Open-LLaVA-Video-R1 \citep{Tang2025LlavaVideoR1} ] 
[\href{https://github.com/Hui-design/Open-LLaVA-Video-R1}{\raisebox{-0.35ex}{\protect\includegraphics[height=1em]{laptop.jpg}}Code}] 

\subsection{Medical Vision}
\textbf{Papers}

[2504] [ChestX-Reasoner \citep{fan2025chestx} ]  
\href{https://arxiv.org/pdf/2504.20930}{ChestX-Reasoner: Advancing Radiology Foundation Models with Reasoning through Step-by-Step Verification} 

[2503] [Med-R1 \citep{lai2025med} ]  
\href{https://arxiv.org/abs/2503.13939v3}{Med-R1: Reinforcement Learning for Generalizable Medical Reasoning in Vision-Language Models} [\href{https://huggingface.co/yuxianglai117/Med-R1}{\raisebox{-0.35ex}{\protect\includegraphics[height=1em]{huggingface.jpg}}Model}] [\href{https://github.com/Yuxiang-Lai117/Med-R1}{\raisebox{-0.35ex}{\protect\includegraphics[height=1em]{laptop.jpg}}Code}]

[2502] [MedVLM-R1 \citep{pan2025medvlm} ] \href{https://arxiv.org/abs/2502.19634}{MedVLM-R1: Incentivizing Medical Reasoning Capability of Vision-Language Models (VLMs) via Reinforcement Learning} [\href{https://huggingface.co/JZPeterPan/MedVLM-R1}{\raisebox{-0.35ex}{\protect\includegraphics[height=1em]{huggingface.jpg}}Model}] 

\subsection{Embodied Vision}
\textbf{Papers}

[2504] [Embodied-R \citep{zhao2025embodied} ] \href{https://arxiv.org/abs/2504.12680}{Embodied-R: Collaborative Framework for Activating Embodied Spatial Reasoning in Foundation Models via Reinforcement Learning} [\href{https://github.com/EmbodiedCity/Embodied-R.code}{\raisebox{-0.35ex}{\protect\includegraphics[height=1em]{laptop.jpg}}Code}]

[2503] [Embodied-Reasoner \citep{liu2025visual} ] \href{https://arxiv.org/abs/2503.21696v1}{Embodied-Reasoner: Synergizing Visual Search, Reasoning, and Action for Embodied Interactive Tasks} 
[\href{https://embodied-reasoner.github.io/}{\raisebox{-0.35ex}{\protect\includegraphics[height=1em]{earth.jpg}}Project}] 
[\href{https://huggingface.co/datasets/zwq2018/embodied_reasoner}{\raisebox{-0.35ex}{\protect\includegraphics[height=1em]{huggingface.jpg}}Dataset}][\href{https://github.com/zwq2018/embodied_reasoner}{\raisebox{-0.35ex}{\protect\includegraphics[height=1em]{laptop.jpg}}Code}]

\subsection{Multimodal Reward Model}
\textbf{Papers}

[2505] [Skywork-VL Reward \citep{wang2025skywork} ] 
\href{https://arxiv.org/abs/2505.07263}{Skywork-VL Reward: An Effective Reward Model for Multimodal Understanding and Reasoning} [\href{https://huggingface.co/Skywork/Skywork-VL-Reward-7B}{\raisebox{-0.35ex}{\protect\includegraphics[height=1em]{huggingface.jpg}}Model}] [\href{https://github.com/SkyworkAI/Skywork-R1V}{\raisebox{-0.35ex}{\protect\includegraphics[height=1em]{laptop.jpg}}Code}]

[2505] [UnifiedReward-Think \citep{wang2025unified} ] 
\href{https://arxiv.org/abs/2505.03318}{Unified Multimodal Chain-of-Thought Reward Model through Reinforcement Fine-Tuning} [\href{https://codegoat24.github.io/UnifiedReward/think}{\raisebox{-0.35ex}{\protect\includegraphics[height=1em]{earth.jpg}}Project}] [\href{https://huggingface.co/collections/CodeGoat24/unifiedreward-models-67c3008148c3a380d15ac63a}{\raisebox{-0.35ex}{\protect\includegraphics[height=1em]{huggingface.jpg}}Models}] [\href{https://huggingface.co/collections/CodeGoat24/unifiedreward-training-data-67c300d4fd5eff00fa7f1ede}{\raisebox{-0.35ex}{\protect\includegraphics[height=1em]{huggingface.jpg}}Datasets}] [\href{https://github.com/CodeGoat24/UnifiedReward}{\raisebox{-0.35ex}{\protect\includegraphics[height=1em]{laptop.jpg}}Code}]

[2505] [R1-Reward \citep{zhang2025r1reward} ] 
\href{https://arxiv.org/abs/2505.02835}{R1-Reward: Training Multimodal Reward Model Through Stable Reinforcement Learning} [\href{https://huggingface.co/yifanzhang114/R1-Reward}{\raisebox{-0.35ex}{\protect\includegraphics[height=1em]{huggingface.jpg}}Model}] [\href{https://huggingface.co/datasets/yifanzhang114/R1-Reward-RL}{\raisebox{-0.35ex}{\protect\includegraphics[height=1em]{huggingface.jpg}}Dataset}] [\href{https://github.com/yfzhang114/r1_reward}{\raisebox{-0.35ex}{\protect\includegraphics[height=1em]{laptop.jpg}}Code}]

\subsection{Audio}
\textbf{Papers}

[2504] [SARI \citep{wen2025sari} ] 
\href{https://arxiv.org/pdf/2504.15900}{SARI: Structured Audio Reasoning via Curriculum-Guided Reinforcement Learning} 

[2503] [R1-AQA \citep{li2025reinforcement} ] 
\href{https://arxiv.org/abs/2503.11197v2}{Reinforcement Learning Outperforms Supervised Fine-Tuning: A Case Study on Audio Question Answering} [\href{https://huggingface.co/mispeech/r1-aqa}{\raisebox{-0.35ex}{\protect\includegraphics[height=1em]{huggingface.jpg}}Model}] [\href{https://github.com/xiaomi-research/r1-aqa}{\raisebox{-0.35ex}{\protect\includegraphics[height=1em]{laptop.jpg}}Code}]

[2503] [Audio-Reasoner \citep{xie2025audio} ] 
\href{https://arxiv.org/abs/2503.02318}{Audio-Reasoner: Improving Reasoning Capability in Large Audio Language Models} [\href{https://xzf-thu.github.io/Audio-Reasoner/}{\raisebox{-0.35ex}{\protect\includegraphics[height=1em]{earth.jpg}}Project}] [\href{https://huggingface.co/zhifeixie/Audio-Reasoner}{\raisebox{-0.35ex}{\protect\includegraphics[height=1em]{huggingface.jpg}}Model}] [\href{https://github.com/xzf-thu/Audio-Reasoner}{\raisebox{-0.35ex}{\protect\includegraphics[height=1em]{laptop.jpg}}Code}]

\subsection{Omni}
\textbf{Papers}

[2505] [EchoInk-R1 \citep{xing2025echoink} ] \href{https://arxiv.org/pdf/2505.04623}{EchoInk-R1: Exploring Audio-Visual Reasoning in Multimodal LLMs via Reinforcement Learning} [\href{https://huggingface.co/harryhsing/EchoInk-R1-7B}{\raisebox{-0.35ex}{\protect\includegraphics[height=1em]{huggingface.jpg}}Model}] [\href{https://huggingface.co/datasets/harryhsing/OmniInstruct_V1_AVQA_R1}{\raisebox{-0.35ex}{\protect\includegraphics[height=1em]{huggingface.jpg}}Dataset}] [\href{https://github.com/HarryHsing/EchoInk}{\raisebox{-0.35ex}{\protect\includegraphics[height=1em]{laptop.jpg}}Code}]

[2503] [R1-Omni \citep{zhao2025r1} ] 
\href{https://arxiv.org/abs/2503.05379}{R1-Omni: Explainable Omni-Multimodal Emotion Recognition with Reinforcement Learning} [\href{https://huggingface.co/StarJiaxing/R1-Omni-0.5B}{\raisebox{-0.35ex}{\protect\includegraphics[height=1em]{huggingface.jpg}}Model}] [\href{https://github.com/HumanMLLM/R1-Omni}{\raisebox{-0.35ex}{\protect\includegraphics[height=1em]{laptop.jpg}}Code}]

\subsection{GUI}
\textbf{Papers}

[2504] [InfiGUI-R1 \citep{liu2025infigui} ] 
\href{https://arxiv.org/abs/2504.14239}{InfiGUI-R1: Advancing Multimodal GUI Agents from Reactive Actors to Deliberative Reasoners} [\href{https://huggingface.co/Reallm-Labs/InfiGUI-R1-3B}{\raisebox{-0.35ex}{\protect\includegraphics[height=1em]{huggingface.jpg}}Model}] [\href{https://github.com/Reallm-Labs/InfiGUI-R1}{\raisebox{-0.35ex}{\protect\includegraphics[height=1em]{laptop.jpg}}Code}]

[2504] [GUI-R1 \citep{lu2025ui} ] \href{https://arxiv.org/abs/2504.10458}{GUI-R1: A Generalist R1-Style Vision-Language Action Model For GUI Agents} [\href{https://huggingface.co/ritzzai/GUI-R1}{\raisebox{-0.35ex}{\protect\includegraphics[height=1em]{huggingface.jpg}}Model}] [\href{https://huggingface.co/datasets/ritzzai/GUI-R1}{\raisebox{-0.35ex}{\protect\includegraphics[height=1em]{huggingface.jpg}}Dataset}] [\href{https://github.com/ritzz-ai/GUI-R1}{\raisebox{-0.35ex}{\protect\includegraphics[height=1em]{laptop.jpg}}Code}]

[2503] [UI-R1 \citep{luo2025gui} ] \href{https://arxiv.org/abs/2503.21620}{UI-R1: Enhancing Action Prediction of GUI Agents by Reinforcement Learning} 

\subsection{Framework}
\textbf{Open-Source Project (Repository without Paper)}

[EasyR1 \citep{zheng2025easyr1} ] 
[\href{https://github.com/hiyouga/EasyR1}{\raisebox{-0.35ex}{\protect\includegraphics[height=1em]{laptop.jpg}}Code}] 

\subsection{Metaverse}
\textbf{Paper}

[2503] [MetaSpatial \citep{pan2025metaspatial} ] \href{https://arxiv.org/abs/2503.18470}{MetaSpatial: Reinforcing 3D Spatial Reasoning in VLMs for the Metaverse} [\href{https://huggingface.co/datasets/zhenyupan/3d_layout_reasoning}{\raisebox{-0.35ex}{\protect\includegraphics[height=1em]{huggingface.jpg}}Dataset}] [\href{https://github.com/PzySeere/MetaSpatial}{\raisebox{-0.35ex}{\protect\includegraphics[height=1em]{laptop.jpg}}Code}]

\subsection{Agents}
\textbf{Open-Source Project (Repository without Paper)}

[VAGEN \citep{VAGEN} ] 
[\href{https://github.com/RAGEN-AI/VAGEN}{\raisebox{-0.35ex}{\protect\includegraphics[height=1em]{laptop.jpg}}Code}] 

\end{document}